\documentclass[sn-mathphys,iicol]{sn-jnl}

\jyear{2022}%

\theoremstyle{thmstyleone}%
\theoremstyle{thmstyletwo}%

\theoremstyle{thmstylethree}%

\begin{document}

\title[Article Title]{Autonomous Mapping and Navigation using Fiducial Markers and Pan-Tilt Camera for Assisting Indoor Mobility of Blind and Visually Impaired People}


\author*[1]{\fnm{Dharmateja} \sur{Adapa}}\email{p20190417@pilani.bits-pilani.ac.in}

\author[1]{\fnm{Virendra Singh} \sur{Shekhawat}}\email{vsshekhawat@pilani.bits-pilani.ac.in}

\author[1]{\fnm{Avinash} \sur{Gautam}}\email{avinash@pilani.bits-pilani.ac.in}

\author[1]{\fnm{Sudeept} \sur{Mohan}}\email{sudeeptm@pilani.bits-pilani.ac.in}

\affil[1]{\orgdiv{Department of Computer Science and Information Systems}, \orgname{Birla Institute of Technology and Science, Pilani}, \orgaddress{\state{Rajasthan}, \country{India}}}

 
\abstract{Large indoor spaces have complex layouts making them difficult to navigate. Indoor spaces in hospitals, universities, shopping complexes, etc., carry multi-modal information in the form of text and symbols. Hence, it is difficult for Blind and Visually Impaired (BVI) people to independently navigate such spaces. Indoor environments are usually GPS-denied; therefore, Bluetooth-based, WiFi-based, or Range-based methods are used for localization. These methods have high setup costs, lesser accuracy, and sometimes need special sensing equipment. We propose a Visual Assist (VA) system for the indoor navigation of BVI individuals using visual Fiducial markers for localization. State-of-the-art (SOTA) approaches for visual localization using Fiducial markers use fixed cameras having a narrow field of view. These approaches stop tracking the markers when they are out of sight. We employ a Pan-Tilt turret-mounted camera which enhances the field of view to 360° for enhanced marker tracking. We, therefore, need fewer markers for mapping and navigation. The efficacy of the proposed VA system is measured on three metrics, i.e., RMSE (Root Mean Square Error), ADNN (Average Distance to Nearest Neighbours), and ATE (Absolute Trajectory Error). Our system outperforms Hector-SLAM, ORB-SLAM3, and UcoSLAM. The proposed system achieves localization accuracy within $\pm8cm$ compared to $\pm12cm$ and $\pm10cm$ for ORB-SLAM3 and UcoSLAM, respectively. 
}

\keywords{Mapping, Localization, Simultaneous Localization and Mapping (SLAM), Fiducial Markers, Indoor Navigation, Pan-Tilt (PT) camera.}



\maketitle

\section{Introduction}\label{sec1}
\noindent Complex Indoor environments pose a variety of challenges in the field of mapping and navigation for Blind and Visually Impaired (BVI) individuals. Visual Assist (VA) systems developed to assist such individuals employ various techniques for improved localization of the user. Efficient mapping in correspondence with accurate localization is used in a majority of such systems providing navigation assistance in complex multi-floor layouts. In addition, rich features present in the indoor environment can be mapped for improving the navigation and localization of the agent. Floor plans and architectural drawings offer incomplete semantic information of the environment \cite{Cheraghi2017}. Tolerances in the dimensions of features in the floor plan may result in inaccurate navigation. In most VA systems designed for indoor way-finding, prior mapping of the environment is performed \cite{Lee2016}.
\\
\noindent Electronic Orientation Aids (EOAs) use tracking and localization information from the environment to assist the users. Indoor environments are GPS-denied as the GPS signal is not accessible in multi-floor layouts, alternatives have been proposed in literature \cite{Plikynas2020}. Cheraghi et al. \cite{Cheraghi2018} use Bluetooth Low Energy (BLE) beacons placed in the environment with a mobile phone estimating the location based on the signal strength of the beacons in its vicinity. The use of BLE beacons eliminates the defacing of the environment but incurs initial as well as running costs. Evennou et al. \cite{evennou2006advanced} attempt to make use of existing WiFi infrastructure commonly present in institutional buildings such as shopping malls, schools, workspaces, and hospitals. WiFi-based localization has not proven to be very accurate and requires uniform WiFi coverage across the building space \cite{He2015}. Mulloni et al. \cite{Mulloni2009} propose the use of artificial features in the form of visual Fiducial markers placed in the environment to achieve indoor positioning and navigation using phone cameras. Visual Fiducial markers are 2D images printed on paper and are cost-effective. Placing a large number of cost-effective visual markers results in defacing the environment and can be an eyesore. We found that few implementations require minimal or inexpensive changes to the environment or use existing infrastructure to provide cues for localization \cite{DelaPuente2019} \cite{Katzschmann2018}.
\\
\noindent Our work proposes a novel VA System for indoor navigation of Blind and Visually Impaired (BVI) individuals that uses visual Fiducial markers for localization. Localization from the visual markers is used for mapping and navigation. During the mapping phase, the markers are placed in the indoor environment within close range of each other. A mobile robot with a Pan-Tilt (PT) turret-mounted camera tracks the markers for localization and uses a 2D laser scanner to construct an occupancy grid map of the environment. A novel marker placement algorithm decomposes the map to generate marker placement candidates. These candidates are pruned to reduce the number of markers required while ensuring complete coverage of the environment. Markers are placed permanently at the locations generated by the placement algorithm. In the navigation phase, the turret-mounted camera atop the mobile robot tracks the reduced markers in the environment, providing continuous localization for directions and path correction. The same turret-mounted camera is mounted on a helmet worn by the user. The user provides voice commands to convey the intended destination. Based on the localization and tracking information, appropriate voice commands are relayed to keep the user on course toward the destination.
\\
\noindent Localization error observed in the experiments conducted was within $ \pm 8 cm $. We observe mapping performance to be better in comparison to existing state-of-the-art methods. Navigation of the robot in the mapped environment functioned smoothly with bounded error within the error of localization.
\\
\noindent The paper is structured as follows: Section \ref{sec:related} describes the current state-of-the-art in this domain. Section \ref{sec:overview} presents an overview of the proposed VA system. Section \ref{sec:methodology} explains the localization, mapping, and navigation process in detail. Section \ref{sec:experiments} documents the experimentation methodology and results obtained from comparisons to the state-of-the-art, followed by the conclusion to the paper.

\section{Related Work}
\label{sec:related}
\noindent VA systems have been in existence since 1964 \cite{kay1964ultrasonic}. Dakopoulos et al. \cite{Dakopoulos2010} classify VA systems based on the level of functionality offered. Electronic Travel Aids (ETAs) convey information about the environment in a different sensory mode. Electronic Orientation Aids (EOAs) provide orientation information to the user during travel, mostly in the form of devices that can be carried by the user. Position Locator Devices (PLDs) \cite{Elmannai2017} incorporate global localization and navigation mechanisms such as GPS. EOAs are commonly used for navigating indoor environments. Indoor navigation systems for BVI individuals rely on novel sources of localization other than GPS while presenting new approaches for tracking and navigation of the BVI user \cite{Silva2017}. Fallah et al. \cite{Fallah2013} enumerate the challenges in developing VA systems as 1) Indoor Localization, 2) Path Planning, 3) representation of the environment, and 4) user interfacing. Many survey papers classify VA systems on this basis. Our focus is on the first three points. 

\subsection{Localization Techniques in Indoor environments}
\label{sec:related_loc}
\begin{table*}
\begin{center}
\caption{Localization techniques used in indoor environments}
\label{tab:va_localization}
\begin{tabular}{|p{0.26\textwidth} | p{0.26\textwidth} | p{0.26\textwidth} | p{0.11\textwidth}|}
\hline
\textbf{Sensor} & \textbf{Techniques} & \textbf{Drawbacks} & \textbf{References}\\
\hline
BLE& BLE signal fingerprinting & Error is up to 2m. & \cite{Guerreiro2019},\cite{Murata2019}\\
 & &Low coverage & \\
\hline
Camera Based& Visual Odometry & Textured walls & \cite{He2015},\cite{Ramesh2018}\\
\hline
LIDAR Based& Scan Matching & Features required in range of LIDAR & \cite{zhang2014loam}\cite{DelaPuente2019}\\
&& Long corridors & \cite{liu2021balm}\\
\hline
Depth Camera Based& Feature matching & Computationally expensive &\cite{Lee2016}\\
&& Requires expensive sensors &\\
\hline
\end{tabular}
\end{center}
\end{table*}
\noindent Localization techniques commonly used in indoor environments are shown in Table \ref{tab:va_localization}. BLE beacons are based on received signal strength, and error can go up to 2 meters. A large number of beacons are required to cover a relatively small environment. Feature mapping-based visual odometry relies on identifiable visual features in the environment derived from RGB images. This method shows signs of failure when there are textured features like wallpapers, floorboards, etc. Scan matching relies on the range data from the sensor to identify features in the environment for associating points received from consecutive scans. Depth cameras provide a rich representation of the environment and, hence require more computation for processing. Depth sensors are relatively expensive compared to the techniques mentioned above.
\\
\noindent Visual Fiducial markers \cite{Garrido-Jurado2016} are artificial markers that can help with localization when seen with a camera. Visual Fiducial markers are binary patterns that can be identified and decoded from an RGB data stream (camera input). They are cost-effective and easy to set up since they can be printed on paper. The range of Fiducial markers is only limited by the resolution of the camera used and the size of the marker. Segmentation of the marker from the surrounding environment eliminates the feature extraction step from the visual odometry pipeline. visual Fiducial markers may deface the environment and may not be convenient to set up everywhere. We present a reduced marker placement approach to ensure minimal use of the visual markers for complete coverage of the given indoor environment.

 \subsection{Mapping Techniques in Indoor environments}
\label{sec:related_map}

\begin{table*}[h]
\begin{center}
\begin{minipage}{\textwidth}
\caption{Mapping techniques used in indoor environments}
\label{tab:va_mapping}
\begin{tabular}{|p{0.26\textwidth} | p{0.26\textwidth} | p{0.26\textwidth} | p{0.11\textwidth}|}
\hline
\textbf{Techniques} & \textbf{Advantages} & \textbf{Drawbacks} & \textbf{References}\\
\hline
CAD Drawings/Manual Map & Easy to implement. Maps boundaries of environment very well & Cannot map all Features. Hard to map new environments &\cite{Lahav2008},\cite{Guerreiro2019}\\
\hline
Record and Play & Easy to map new Environments & Data storage requirements &\cite{Bousbia-Salah2011}\\
\hline
BLE signal Fingerprinting & Less defacing/easily embedded & Require a base map &\cite{Murata2019}\\
\hline
Visual Odometry & Single Camera and IMU & Sensitive to abrupt motion &\cite{He2015},\cite{Ramesh2018}\\
\hline
Feature Matching & Generates detailed map. Semantic Extensions.  & Computationally expensive. Bulky equipment required &\cite{Lee2016},\cite{DelaPuente2019}\\
\hline
\end{tabular}

\end{minipage}
\end{center}
\end{table*}

The majority of VA systems are designed to be Electronic Travel Aids (ETAs) \cite{Tapu2018} aimed at obstacle detection on the path of the BVI user. ETAs work by understanding the environment through various sensors and conveying the features observed to the user by various sensory inputs (audio, haptics, etc.). Mapping of the environment is not required in ETAs. Hence, they are excluded from the mapping-based classification of VA systems. Electronic Orientation Aids (EOAs) require mapping and environment representation to help with localization and navigation of the user. Mapping techniques used in EOAs found in the literature are listed in Table \ref{tab:va_mapping}. Some systems implement live mapping, whereas others use a saved map representation of the known environment. Live mapping is favoured in the case of systems using Visual Odometry \cite{Gutiérrez-Gómez2012} and SLAM \cite{Ramesh2018},\cite{zhang2014loam},\cite{DelaPuente2019},\cite{liu2021balm}. Ishihara et al. \cite{ishihara2017beacon} propose a hybrid system that incorporates Structure-From-Motion (SFM) in conjunction with BLE-based localization corrections. Live mapping of the environment involves complex computation, and requires more computing resources. Adding computing requirements to the VA system increases the bulk of the wearable device.  Creating floorplans and layouts for map information is time-consuming. Static maps of the known environment favour layouts that remain static over a long time. However, the use of known maps reduces the bulk of computing resources. Hence, we propose a VA system that separates the mapping step from navigation, leaving minimal equipment to be mounted on the wearable device.

\subsection{Visual SLAM Methods}
Visual Simultaneous Location and Mapping (vSLAM) use visual features from the environment captured by an RGB camera for the purpose of SLAM. This idea was first presented by Davison et al. \cite{davison2007monoslam} and Eade and Drummond \cite{eade2006scalable}. Based on the type of landmarks used for SLAM, vSLAM methods can be classified into natural feature-based SLAM or artificial feature-based SLAM. Natural features include visual features present in the environment that do not change over time, for example, doors, fire extinguishers, benches, etc. Artificial landmarks are features placed in the environment specifically for the purpose of enhancing mapping, localization, or both. e.g., BLE, visual markers. The use of visual markers can also be termed as marker-based SLAM. The following sections provide further details on these two types of vSLAM.
\subsubsection{Natural Feature-based SLAM}
Natural feature-based SLAM attempts to utilize existing image features in the environment. Most of the methods vary based on their approach to identifying or tracking these features. PTAM (Parallel Tracking And Mapping) \cite{klein2007parallel} utilizes FAST (Features from Accelerated Segment Test) feature corner detector for establishing key points that are matched by patch correlation. The implementation of PTAM splits tracking and mapping into two separate threads. This idea led to an increase in the real-time performance of subsequent methods. PTAM suffers from its inability to properly relocalize.
Mur-Artal et al. \cite{mur2015orb} developed ORB-SLAM, a SLAM system using the ORB (Oriented FAST and Rotated BRIEF) feature detector. These ORB key points are used for tracking and re-localization using a Bag of Words (BoW) approach to address the re-localization problem. The ORB-SLAM system estimates the camera position and the structure of the environment by reducing the reprojection error of the key points across known keyframes. This system is improved in ORB-SLAM2 \cite{mur2017orb}, extending its capability to work with stereo and RGBD cameras. Recently, the system has been extended to ORB-SLAM3 \cite{campos2021orb} which includes inertial information, as well as the creation of multiple maps, which improves the process of re-localization.
Feature-based SLAM has several drawbacks. Firstly, re-localization is not reliable (sometimes impossible) in an environment with repetitive structures. Monocular pose estimation of key points results in the scale of the map being drifted or skewed. Hence, when using monocular cameras only, the scale of the generated map is unknown. Key points in the environment considered as features for these methods pose an inherent challenge in environments/scenes with low texture. Hence, these methods require some amount of texture in order to extract reliable features. It is simply impossible in some indoor environments with large white walls and ceilings. Finally, many of the key points employed for tracking and re-localization correspond to short-term features such as corners in shadows or objects that are transient.

\subsubsection{Marker-based SLAM}
MarkerMapper \cite{munoz2018mapping} proposes the use of ArUCO markers for mapping indoor environments. 
\noindent The SPM-SLAM (Squared Planar Marker based SLAM) \cite{Munoz-Salinas2019} system proposes a method for creating a map of planar markers and localizing the camera pose from a video sequence. In addition, the method deals with the ambiguity problem produced by the use of planar markers, avoiding incorrect camera pose estimation during the SLAM process. SPM-SLAM requires at least two markers in each frame in order to build a map. To reduce the dependability on markers, UcoSLAM \cite{Munoz-Salinas2020} proposes combining key points and square Fiducial markers placed in the environment. Fiducial markers act as stable features in the environment over time, with image features acting as temporary or short-term features, enabling UcoSLAM to perform robust tracking in most scenarios. UcoSLAM takes advantage of the benefits offered by Fiducial markers as soon as one is detected in the scene, allowing it to obtain maps with a correct scale. In addition, the system maintains a set of map points of the environment, which is updated as new frames are added to the system. The map points provide continuous tracking even in large scenes, and they take part in both the initialization and the optimization of the map. The main drawback of feature-based methods is that the number of map points increases as the environment grows, increasing the computation time. On the other hand, many map points were created by key points in areas that tend to change over time. In sSLAM \cite{Romero-Ramirez2023}, the authors propose a new system based on key points and Fiducial markers. sSLAM also uses a map of markers created in previous frames. Then, the system reuses the map by relying on scene Fiducial markers and key points. The key points are used temporarily only for continuous tracking.

\begin{figure}[!t]
\centering
\includegraphics[width=\columnwidth]{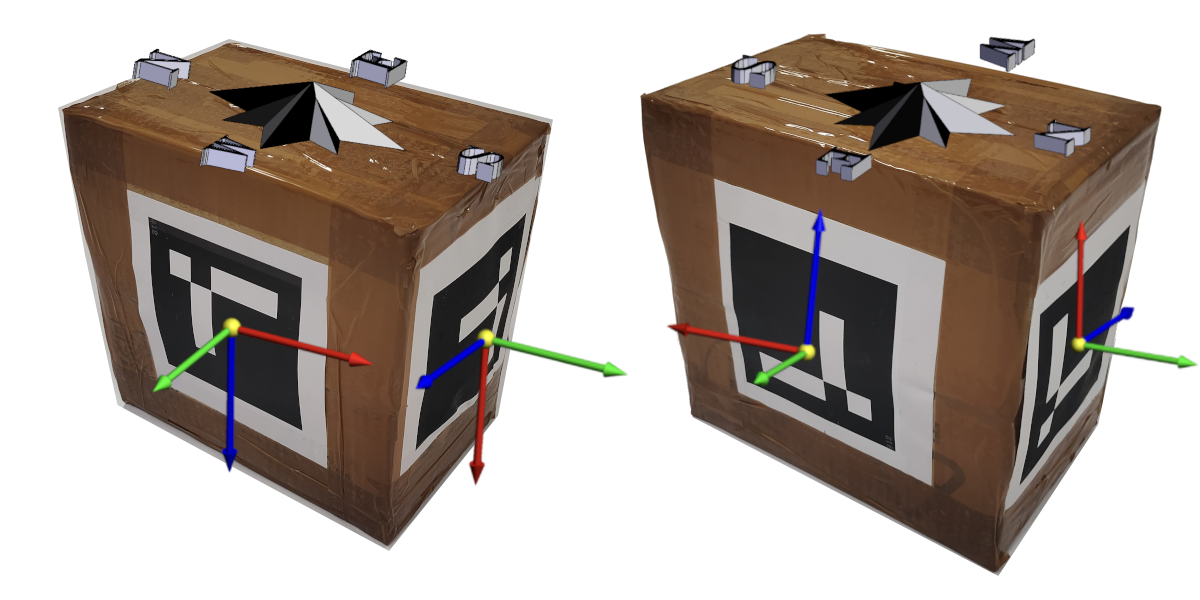}
\caption{Cuboid with the marker in different orientations on 4 sides. Green, red, and blue arrows represent the x,y, and z axis, respectively}
\label{fig:360_marker}
\end{figure}

We can observe that the utility of keypoints is gradually reducing given these two SLAM systems. Relying on key points is also not favourable in texture-less environments such as large rooms, corridors, etc. There is also the issue of scale in environments mapped using features alone, reducing the usability of the map for navigation. There is also a dependency on the resolution of the camera to clearly identify and track features at larger distances. Increasing the number of features may solve this to some extent but will worsen the computation time.
\\
\noindent In our approach, we are relying solely on marker-based localization. With a marker and feature-based localization, some sections of the environment can be left without markers for the feature detector to take over. In the case of a marker-only approach, we need to place markers in a way that all points in the environment should lie within the range of at least one marker. To address this, we propose the use of cuboid-shaped features with planar markers on four sides of the cuboid \ref{fig:360_marker}. This ensures that the presence of one cuboid enables 360° coverage of its vicinity. In addition, we also propose a reduced marker placement algorithm for reducing the number of markers required to cover a mapped indoor environment. The advantages of a marker-based approach for localization are (1) Localization is as accurate as state-of-the-art methods and can be implemented in real-time, (2) Computation time does not scale with the increasing map size as is the case with feature-based approaches. (3) Scaling and resolution problems are eliminated due to the use of a defined set of visual markers.

\subsection{State of the art VA systems}
\label{sec:related_sota}
\noindent Cheraghi et al. \cite{Cheraghi2017} \cite{Cheraghi2018} suggest the use of multiple proximity sensors in the form of Bluetooth Low Energy (BLE) beacons scattered across the indoor environment for localizing for guiding a BVI user. The user interacts with a smartphone-based system that localizes based on proximity detection from multiple beacons in the vicinity. Voice input is taken from the user regarding the destination, and the navigation instructions are given in the form of voice prompts or tactical feedback from the mobile phone. Experiments were conducted with both sighted and BVI users, and the effectiveness of the system was adjudged based on feedback from the users and performance measures such as navigation time and navigation distance.
\\
\noindent In Navcog3 \cite{Sato2019}, the authors have considered including semantic information to assist the goals of the system. They have stated that semantic information is useful for understanding the environment as well as for improving the localization of the user. The localization information is derived using BLE beacons. To improve accuracy, they have used a large number of beacons (in the order of 220 beacons over a $21000 m^{2}$ area) scattered across the environment. The battery-powered Bluetooth beacons last for 1 year, and costs go up with the scale of the environment. An exhaustive list of landmarks and semantic features have been specified for improved traversal of multi-level layouts. They have also supported the use of a separate system for conducting one-time mapping of the environment.
\\
\noindent Lee et al. \cite{Lee2016} propose a novel wearable RGBD camera-based indoor navigation system for the visually impaired. The wearable device consists of a jacket containing a laptop running the navigation software, a MEMS (Micro-Electro-Mechanical System) IMU (Inertial Measurement Unit), and a haptic-based interface for conveying directions to the user. A head-mounted RGBD camera connects to the laptop, and the entire system is connected to a smartphone for receiving and conveying user input to the system. The navigation software consists of a pose estimation algorithm, a mapping algorithm, and a dynamic path planning algorithm. Experiments were conducted to verify each component of the system as well as its entirety. The first component verifies the accuracy of the pose estimation by having the subject travel a closed loop in the environment and measuring the estimated localization error at the start point and the end point – which is ideally 0. The proposed algorithm reduces drift from 3.6 meters to 2.6 meters over native FOVIS \cite{Huang2017}. The entire system, armed with prior knowledge of the indoor 2D map and points of interest, is compared to a white cane user with known and unknown regions. The performance of this system is attributed to reduced travel time on average compared to a traditional white cane user with prior knowledge of the environment. The system is able to achieve real-time performance (28.6Hz) given that the RGBD information is available at a resolution of 320x240 pixels.
\\
\noindent He et al. \cite{He2015} propose a wearable device for tracking users in indoor environments using visual-inertial sensors. The accelerometer, magnetometer, and gyroscope on Google Glass are used to perform real-time human motion tracking using an adaptive gain complementary filter. A Gauss-Newton method \cite{Tian2013} is used to achieve this. Correspondence between consecutive images is determined using SURF (Speeded Up Robust Features) \cite{BAY2008346}. The identified feature points observed and tracked in a sequence of consecutive images are stored as landmarks. An additional visual sanity check is put in place such that dynamic points in the environment do not disturb the visual odometry. The system is susceptible to errors caused by abrupt motion (running, sudden turns, etc.) due to the limited processing frequency of the pipeline. Another major limitation is that the tracked trajectory is relative to the starting point, and the scale is not accurate. The authors have suggested future developments regarding the integration of ambient and ego-motion tracking with motion alerts.

\subsection{Our Contribution}
\noindent The state-of-the-art methods focus on improving localization and mapping in indoor environments. There is limited work on using these methods for the navigation of BVI users in unknown indoor environments. Our contribution lies in the use of visual markers and PT turret-mounted camera for localization and navigation in indoor environments.
\begin{itemize}
  \item We design a localization system that uses 360° visual markers in the form of a cuboid. The cuboid faces are markers in different orientations representing different directions faced by the marker in the global map (Figure \ref{fig:360_marker}). The markers are identified by a PT turret-mounted camera for improving the markers' Field of View (FOV). We present an empirical evaluation of marker size for using the system as an effective source of localization.
  \item We map the indoor environment based on localization information from visual markers and range information from a laser scanner.
  \item We present a technique for marker placement reduction. It reduces the number of markers required to completely cover a given environment.
  \item We develop a method for navigating the indoor environment using the software control stack for tracking and transitioning between markers. This system is implemented on a mobile robot platform for verification.
\end{itemize}

\noindent The work presented in this paper contrasts with existing state-of-the-art methods in the following ways:
\begin{itemize}
    \item The localization system uses 360° markers instead of simple planar markers used in the state-of-the-art methods.
    \item The camera is mounted on a PT turret to extend the FOV and localization range. Existing methods use a static-mounted camera.
    \item Localization requires only one marker to be in the range of the camera.
    \item Our system suggests reduced placement of the visual markers given a map of the environment, thus, reducing the number of markers needed during run time.
\end{itemize}

\section{System Overview}
\label{sec:overview}
\begin{figure*}[!t]
\centering
\includegraphics[width=\textwidth]{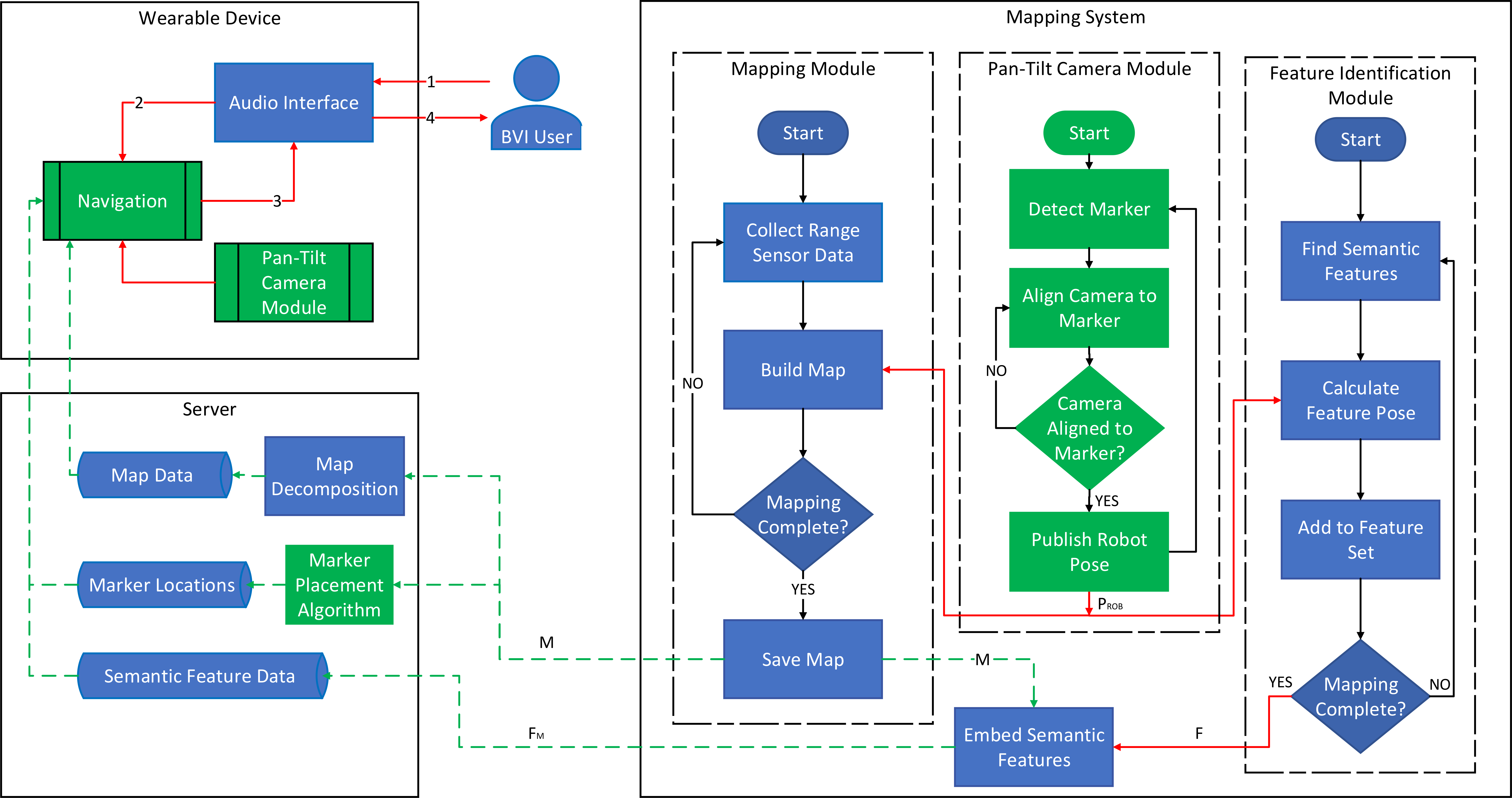}
\caption{Conceptual diagram of the proposed VA System for BVI people. Green-colored blocks indicate the components described in this paper. Black arrows indicate control flow, red arrows indicate continuous data flow, and green arrows indicate intermittent data flow. $P_{ROB}$ conveys the pose of the robot derived from the PT camera module. $M$ represents the compact map representation. $F$ represents the features identified in the environment during mapping. $F_{M}$ indicates the detected features embedded in the map $M$}
\label{fig:dharma_phd}
\end{figure*}

\noindent An overview of the proposed VA system is given in Figure \ref{fig:dharma_phd}. It primarily consists of a mapping system, a server, and a wearable device that will interact with the BVI user for indoor navigation.

\noindent The mapping system is implemented on a mobile robot that consists of a PT turret-mounted on top of it along with a laser scanner for mapping. There are 3 hardware/software modules that are part of the mapping system -  the PT camera module for the localization of the robot, the mapping module for constructing a metric map of the environment, and the feature identification module for identifying semantic features in the environment. The metric map and semantic features identified by the mapping system are saved to the server, where the map is decomposed, and marker locations are generated. The Wearable device is used to interact with the BVI user and guide the user in the mapped indoor space.
\\
\subsection{Mapping System}
\noindent The mapping system constructs a metric map of the environment to identify navigable areas in the indoor spaces. The construction of a map is critical to path planning and navigation. Conducting manual mapping of large indoor spaces is tedious and requires human effort. Hence, we use a mobile robot with the necessary sensors mounted on it to conduct mapping. The robot houses a laser scanner and a PT camera module for range data and localization data, respectively. It also contains a mini PC that runs the necessary software to drive the robot as well as to receive the information from the sensors and build a map out of it. The mapping system contains the following modules:
\begin{enumerate}
    \item PT Camera Module: This module has a camera mounted on a PT turret mechanism, with the turret being controlled by programmable servo motors. The camera mounted on top of the robot detects the visual marker placed in its vicinity and estimates its pose. Given the camera parameters and the size of the markers, perspective transformation is applied to the camera image to localize and identify the visual marker. To further extend the field-of-view of the camera and the sensor, the PT turret enables moving the camera such that the camera stays aligned to the marker for continuous localization. Considering that the markers are static and the camera is mounted on a mobile robot on a PT turret, we can localize the robot with respect to the marker and hence, the indoor space. Only when the marker is detected, and the camera aligned we calculate and publish the pose of the robot to be used by other modules. This module is further elaborated in Section \ref{sec:marker_tracker}.
    \item Mapping Module: The range-sensing laser scanner mounted on the mobile robot is limited by its range and the current location of the robot. Hence, we implement a procedure to stitch the laser scanner data with the localization data from the markers and PT camera module. After obtaining the laser scanner data, the ranges are drawn on a temporary local map. When stable localization information (robot pose) is available from the PT camera module, this local map is transformed to align with the current location and orientation and stitched to a global map. This process is continued till the map completion is manually triggered. The procedure is detailed in Section \ref{sec:env_mapping}.
    \item Feature Identification Module: Indoor spaces contain rich textual or symbolic information associated with sections of the space. This information can range from a  label on a door indicating the room number to fire extinguishers containing operating instructions. This is difficult for a BVI individual to identify and act upon. Some of this information can help improve the scope of navigation in the environment. To capture this information and use it for navigation, we propose a feature identification module that identifies, localizes, and embeds these features in the indoor environment map created. Semantic features considered include door numbers in office spaces, symbols indicating restrooms, water fountains, fire extinguishers, text signs, etc. These features are identified from the camera images captured during mapping. Once identified, the features are localized in the map based on the stable localization from the PT camera module (robot pose) and added to the feature set. When the map completion is manually triggered, all the feature set data is embedded onto the map and saved to the server. When the final marker locations are available after the marker placement algorithm, the features are associated with their nearest marker for better navigation.

\end{enumerate}
The mapping process requires an operator who will place the markers in the environment, control the mapping robot and also indicate completion of mapping. The mapping process is as follows:
\begin{itemize}
    \item The operator places markers at a roughly uniform distance from each other in the indoor environment to be mapped.
    \item The operator then places the mapping robot in the environment and switches it on.
    \item The operator remotely operates the robot to cover the navigable sections of the environment.
    \item On completion of mapping, the operator signals the robot and ensures all mapping data is saved to the server.
\end{itemize}

\subsection{Functioning of Server}
\noindent The server is used to store and relay data to the wearable device for conducting the navigation. The server receives map data with embedded feature data from the mapping robot on completion of mapping. The server runs utility programs for map decomposition and marker placement on the input map data. Map decomposition creates a sparse representation of the map in a graph format from the map received in matrix format (occupancy grid). The marker placement algorithm works on the input map and the marker range to suggest a set of marker locations that ensure 100\% coverage of the map with fewer markers than what was used during mapping. The sparse map representation and suggested marker locations are stored in the server along with the embedded semantic feature information required for the wearable device to conduct navigation. When the wearable device is activated/switched on, it requests the map data, marker locations, and semantic feature data pertaining to the environment from the server.

\subsection{Wearable Device}
The wearable device is in the form of a jacket containing all the necessary computing, power, and user interface systems. The PT camera module is carried over from the mobile robot used for mapping and is mounted on a helmet worn by the user. On startup, the wearable device requests information of the current space (map and features) and the marker locations from the server. The interaction of the wearable device with the user is as follows: 
\begin{enumerate}
    \item The BVI user is prompted via audio to suggest a destination. The user conveys the intended destination to the system using commands like:``Take me to \lt destination\gt ''or ``Go to \lt destination\gt ''.
    \item The navigation module receives the audio input and decodes it to text and keywords. The keywords are then searched in the semantic feature database, and the best result is conveyed to the user for confirmation. After confirmation, the current user location is obtained from the PT camera module, and a path is planned for the user. In case the mapping data does not contain any match for the queried destination, the system conveys an ``Invalid Destination'' response to the user.
    \item The user is guided upon the planned path based on the current user location from the PT Camera module and the waypoints denoting the path. Navigation commands are triggered to keep the user on the planned path and guide the user at intersections.
    \item The audio interface converts the navigation commands to their audio equivalents to be conveyed to the user.
\end{enumerate}
Steps 3 and 4 are repeated till the user reaches the destination. Step 1 is repeated again to initiate navigation to a new destination.
\\

\noindent We will now describe in detail the working of this VA system as an indoor navigation system.

\section{Indoor Mapping and Navigation System}
\label{sec:methodology}
\noindent The complete pipeline of the VA system consists of two stages: mapping and navigation. The mapping phase gathers information about the environment and creates a simple and useful representation for navigation. The navigation phase interfaces with the user to set destinations, plan the path, and guide the user along the path to reach the destination. The system is explained in the context of these two phases.
\subsection{Mapping Phase}
\begin{figure}[!t]
\centering
\includegraphics[width=\columnwidth]{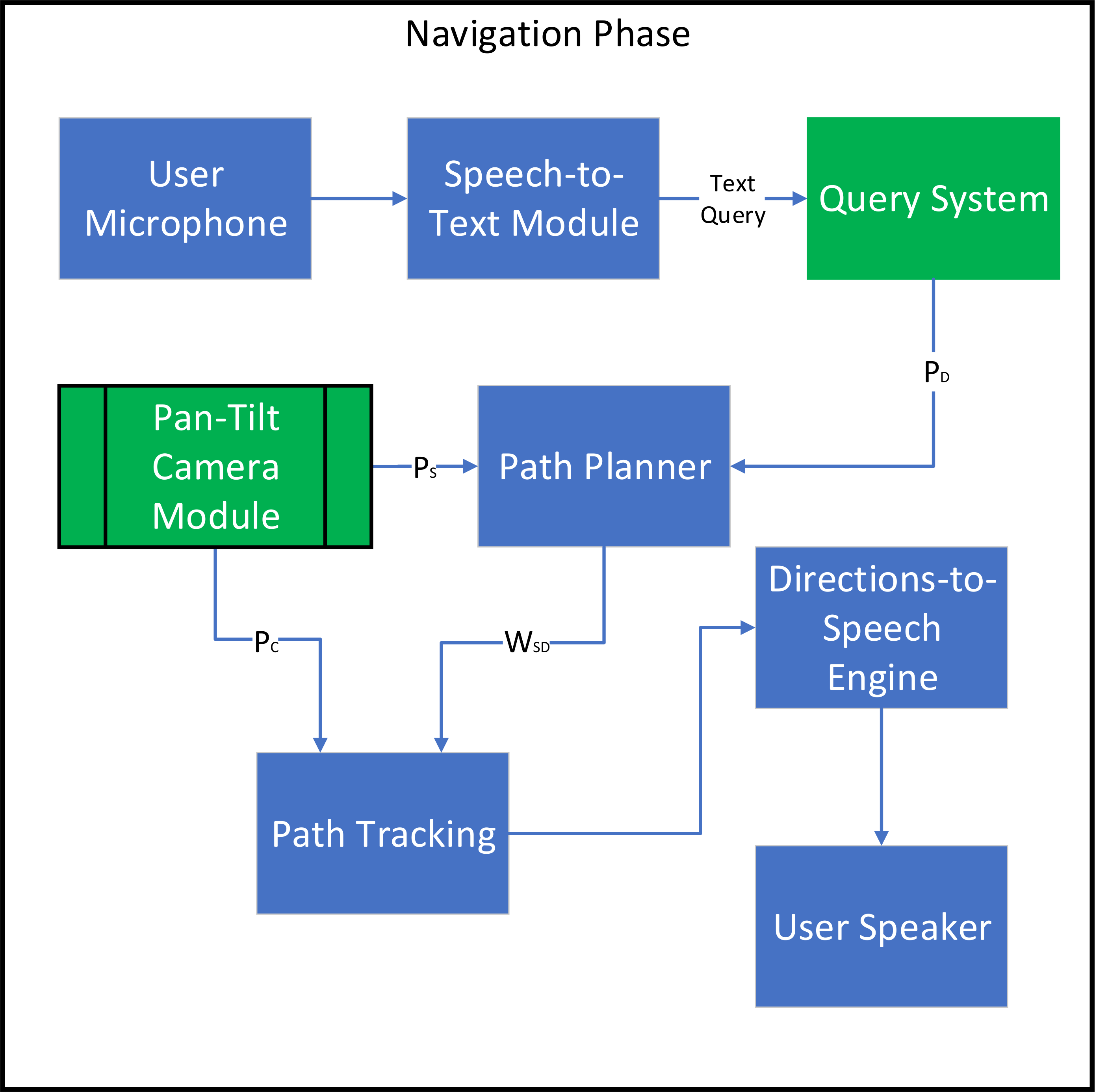}
\caption{High-level block diagram of the mapping system. $P_{ROB}$ conveys the pose of the robot derived from the PT camera module. $M_{O}$ represents the occupancy grid map representation. $M$ represents the compact map representation. $M_{R}$ indicates the marker positions derived after the reduced marker placement. $F$ represents the features identified in the environment during mapping}
\label{fig:Mapping_BD}
\end{figure}
\noindent The mapping phase is performed by the operator, manually controlling the motion of the mapping robot across the environment (Figure \ref{fig:Mapping_BD}). The process is as follows:
\begin{enumerate}
\item The operator places Visual Markers at a uniform spacing with each other within the traversable areas of the environment.
\item The robot position is initialized to global origin (0,0), and the nearest marker is tracked by the marker-based localization system.
\item Information from the localization system and the range scanner is used to construct the map of the environment in the immediate vicinity of the robot.
\item After visualizing the initial environment, the operator starts issuing motion commands to the robot. When the marker being tracked initially is out of reach, the system starts looking for and tracks the next marker after doing a horizontal 360° scoop of the PT turret.
This next marker now provides localization for subsequent mapping.
\item The steps 3-6 are repeated till all traversable areas of the environment are mapped.
\end{enumerate}
Once the mapping is complete, the occupancy grid map is sent to the Voronoi Partitioning, Reduced Marker Placement, and Feature Identification modules.
\begin{enumerate}
\item The Voronoi Partitioning module decomposes the map into a sparse representation for the path planner.
\item Reduced marker placement generates marker positions that cover the entire environment with a smaller number of markers compared to what was used during mapping. This component is present on the server and is run after the mapping has been completed.
\item Feature Identification module identifies semantic features in the environment and embeds them onto the map generated for better navigation capabilities. The features identified are attached to the nearest bounding cell (occupied cell) and stored in the database on the server.
\end{enumerate}

\subsubsection{Visual Fiducial Markers}
\noindent Fiducial markers are artificially placed objects in the environment that assist in the localization of an agent by providing a relative location on identification. Visual Fiducial markers \cite{Kalaitzakis2021} are unique binary patterns coded in the form of black and white rectangles in a grid pattern with an external black border surrounded by white space. When a marker is within viewing range of the camera, the decoded pattern indicates a marker ID number that can be associated with a location in the environment relative to the camera. The calculation of camera localization is based on perspective transforms knowledge of camera parameters and marker size.
\\
\noindent Kalaitzakis et.al. \cite{Kalaitzakis2021} have presented an empirical evaluation of four recently developed visual Fiducial markers. Among these, we have used ArUco Markers \cite{Romero-Ramirez2018} as they have good position and orientation results, good detection rate, and low computation cost for single markers. The disadvantages mentioned specify sensitivity to smaller marker sizes and higher computation costs for larger dictionary size. These are overcome by empirical evaluation of marker ranges (in section \ref{section:marker_range}) and the use of a limited dictionary by reducing the number of markers (in section \ref{sec:reduced_markers}), respectively. We also use a custom library of ArUco markers to further reduce the computing time.
\\
\noindent ArUco markers provide visibility of 160°. To ensure minimal defacing of the environment, we considered localization sources that have a 360° FOV. In this regard, we created a cuboid with markers mounted on the four sides. This will result in one cuboid serving the entire range of the marker all around. An example is shown in Figure \ref{fig:360_marker}. 360° markers are especially useful in intersections where multiple paths intersect and require all-around localization. In case of long corridors, markers are placed only on two opposite sides of the cuboid.

\subsubsection{Marker Pose from ArUco}
The ArUco library \cite{Romero-Ramirez2018} used for detection of the visual Fiducial markers returns a 6-dimensional vector representing the 3 axes of the translation and rotation components, respectively. The image frame acquired from the camera image is processed by the library, and a list of Markers detected in the current frame is returned along with their corresponding translation and rotation vectors. The working of the ArUco Library is as follows:
\begin{enumerate}
\item Image Resize: A downsized version of the input image is used for segmentation so as to reduce computation. It also helps detect various sizes of markers by processing multiple copies of the image at different scales. The resize parameter is dynamically updated after a marker is detected to ensure the marker is tracked even if the camera is in motion.
\item Local Adaptive segmentation: If no markers are detected previously, random global thresholding is performed as per \cite{Romero-Ramirez2018}. Once a marker is detected, Otsu's algorithm \cite{Otsu1979} is employed to select the optimal threshold to adjust to changes in illumination.
\item Contour Extraction: Contours are extracted from the thresholded image using Suzuki and Abe's Algorithm \cite{SUZUKI198532}. The extracted contours undergo polygon approximation, and only convex polygons with four corners are considered to be markers.
\item Marker Code Extraction: From the marker polygon, the binary code is extracted using Otsu's method for binarization. The order of the binary pattern is matched with those available in the dictionary, and the unique ID of the marker is derived.
\item Corner Upsampling: The pose estimate of the marker is based on accurate localization of the corners of the marker in the image. For improved precision in corner detection, upsampled versions of the image segment around the marker's corners undergo corner refinement, and the best location is selected.
\item Pose Estimation: Calibration parameters of the camera combined with affine transforms can estimate the 3D pose of objects in the image. Corners and marker boundaries detected in the previous step, along with the parameters, are used to estimate the 3D pose of the marker given in the form of a 6D vector:
\end{enumerate}
\begin{equation}
\label{eqn:marker_pose}
 P_{marker}=\{T=\{T_x,T_y,T_z\},R=\{\theta_x,\theta_y,\theta_z\}\}
\end{equation}
\noindent This information is published to the ROS topic /turret/pose\_raw, in the form of a custom ROS message. This uses the built-in message type geometry\_msgs/PoseStamped\footnote{\url{http://docs.ros.org/en/kinetic/api/geometry_msgs/html/msg/PoseStamped.html}} to represent the pose estimate of the marker with respect to the turret camera along with an integer representing the unique ID of the markers detected.

\subsubsection{Geometric Transforms}
\begin{figure}[!t]
\centering
\includegraphics[width=\columnwidth]{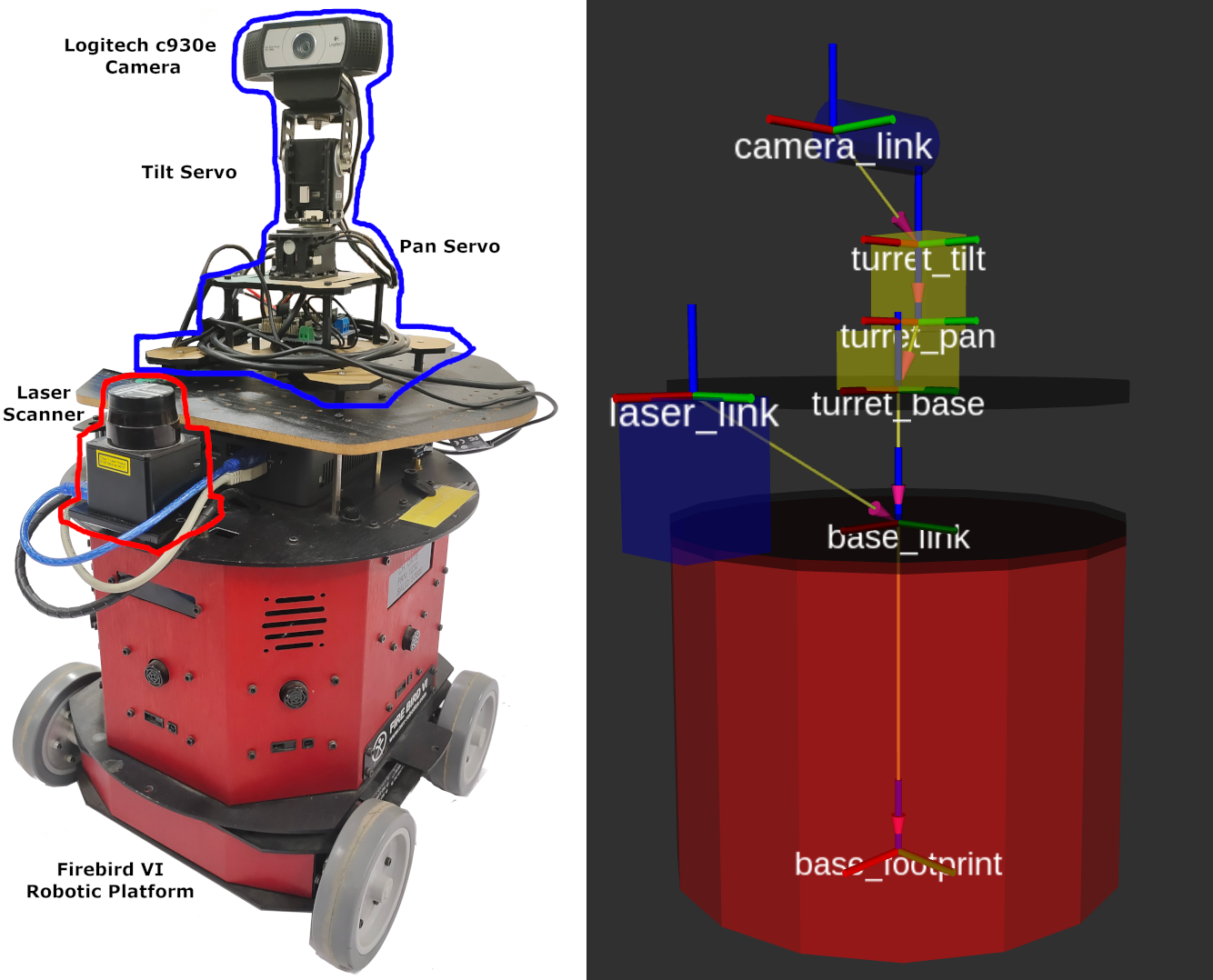}
\caption{Mobile robot configuration for environment mapping (left) and transform tree (right)}
\label{fig:mapping_robot}
\end{figure}

This section explains the geometric transformations required between different frames of reference in the environment corresponding to the map, robot, and the sensors/actuators mounted on the robot. Transforms are handled by the \textit{tf} (transforms) package in ROS\footnote{\url{http://wiki.ros.org/tf}}. The mobile robot setup used for mapping is shown in Figure \ref{fig:mapping_robot}. For the given setup, the following transforms are required for mapping:
\begin{enumerate}
    \item \textbf{Laser Scanner frame to Turret base frame}: The laser scanner is mounted firmly to the robot base.  Hence, a static transformation between the position and orientation of the laser scanner with respect to the turret base. The laser scanner is mounted to the front end of the robot. Hence the transformation will be a translation in the negative y-direction (Equation \ref{eqn:trans_lt}).
    \\
    \begin{equation}
    \label{eqn:trans_lt}
    T_{lt}=
    \begin{bmatrix}
        1 & 0 & 0 & 0 \\
        0 & 1 & 0 & -t_{1}^{y} \\
        0 & 0 & 1 & 0 \\
        0 & 0 & 0 & 1
    \end{bmatrix}
    \end{equation}
    \item\textbf{Camera frame to Turret base}: Since the turret has two degrees of freedom - pan and tilt - The camera frame also has two degrees of freedom in the form of rotation along the z-axis and x-axis. Using the pan and tilt angles ($\theta_{p}$ and $\theta_{t}$) received from the turret controller, we obtain the orientation of the camera frame with respect to the turret base. The composite transform of rotation along the z-axis and rotation along the y-axis is given in Equation \ref{eqn:trans_ct}.
    \\
    \begin{equation}
    \label{eqn:trans_ct}
    T_{ct}=
    \begin{bmatrix}
        \cos\theta_{p} & -\sin\theta_{p}\cos\theta_{t} & \sin\theta_{p}\sin\theta_{t} & 0 \\
        \sin\theta_{p} & \cos\theta_{p}\cos\theta_{t} & -\cos\theta_{p}\sin\theta & 0 \\
        0 & \sin\theta_{p} & \cos\theta_{t} & 0 \\
        0 & 0 & 0 & 1    
    \end{bmatrix}
    \end{equation}
    \item\textbf{Marker frame to Camera frame}: Given the camera image, the ArUco library returns the relative position and orientation vectors of the marker (Equation \ref{eqn:marker_pose}). These vectors are used as is to represent the required transform.
    \item\textbf{Marker frame to Global frame}: Finally,  we need a composite transform to place the marker in the global frame. The first marker detected by the system during mapping is considered to be at the global origin. Subsequent markers are placed according to their relative location to the first marker in the global frame. Since the markers are static, this set of transforms remains static. This is explained in Section \ref{sec:marker_trans}.
\end{enumerate}

The mobile robot operation during navigation differs as mapping and marker placements are already fixed; the marker to global frame transform is now a marker position database. The marker position database specifies the position and orientation of the markers in the global frame based on their IDs and orientation.

\subsubsection{Marker Tracker}
\label{sec:marker_tracker}
\noindent The marker tracker is designed to identify and track the visual markers in the environment. Since the system has a hard dependency on the markers for localization, at least one marker needs to be in the FOV of the camera \cite{Xing_2020}. Even though the markers are designed to be detected from all sides, the limited FOV of the camera increases the number of markers required to satisfy the above criteria. Hence, we developed a PT-enabled turret to add two degrees of freedom to the camera. This ensures that the camera also has an effective horizontal FOV of 360° along with improved vertical FOV. The Marker Tracker software and hardware working is given below.

\begin{itemize}
\item \textbf{Software Implementation} The marker tracker software serves two purposes: (1) derive the relative pose of the robot based on the pose of the marker detected and (2) calculate the pan and tilt angles required to keep the detected marker in view of the camera. The camera image is obtained using OpenCV VideoCapture. It is operated at a resolution of 1920x1080 at 30 fps - the maximum supported by the Logitech c930 camera. The ArUco marker library \cite{Romero-Ramirez2018} is used for segmenting the markers from the video feed. The software works on the principle of perspective projection, given prior knowledge of the camera parameters. Image frames input to the library returns the relative pose of the markers detected. Since the position of the marker in the environment is known, simple transformations are applied to the relative pose obtained from the ArUco Marker library to obtain localization information of the turret.
\\
\begin{figure}[!t]
\centering
\includegraphics[width=\columnwidth]{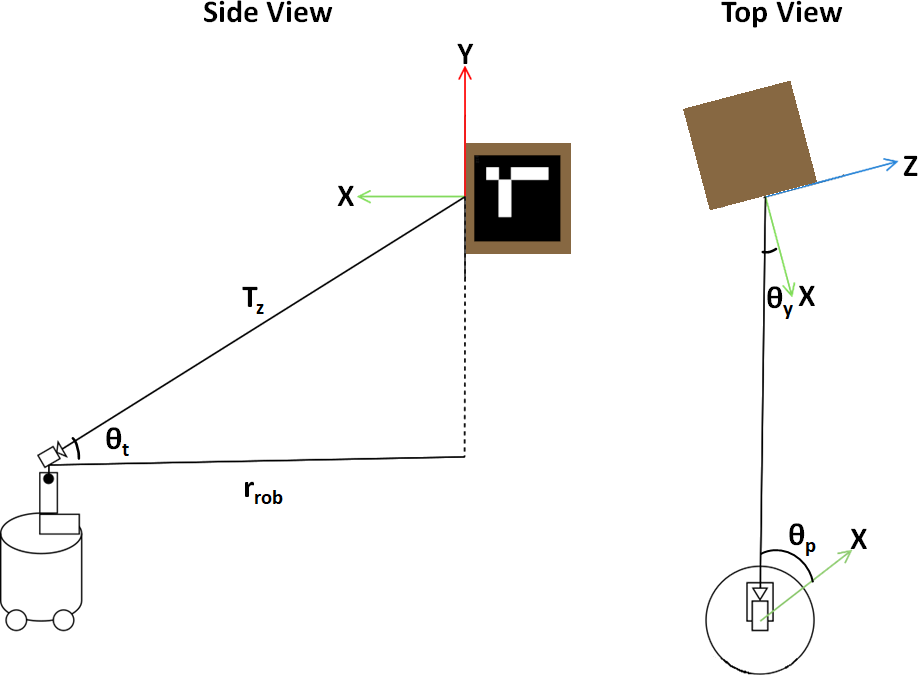}
\caption{Marker tracker geometry for conversion to 2D polar coordinates}
\label{fig:marker_tracker}
\end{figure}
\\
\noindent The relative pose obtained from the ArUco library is given in Equation \ref{eqn:marker_pose}. From this equation, the $T_z$ component specifies the distance from the camera to the marker (refer Figure \ref{fig:marker_tracker}). This is multiplied by the cosine of the tilt angle of the turret to project the distance onto a 2D plane lying along the camera frame \ref{eqn:polar_r}. The rotation along the y-axis (vertical direction) $\theta_y$ of the marker specifies the alignment of the marker with respect to the camera. In addition to that, the pan angle of the turret $\theta_{p}$ helps put into perspective the angle offset between the turret base frame and the marker frame. Hence, we now have location information in the form of polar coordinates as given in Equations \ref{eqn:polar_pose}, \ref{eqn:polar_r}, and \ref{eqn:polar_theta}.
\begin{equation}
\label{eqn:polar_pose}
 p_{rob} = (r_{rob},\theta_{rob})
\end{equation}
where,
\begin{equation}
\label{eqn:polar_r}
 r_{rob} = T_z * cos(\theta_{t})
\end{equation}
\begin{equation}
\label{eqn:polar_theta}
 \theta_{rob} = \theta_{p}
\end{equation}
The polar coordinates are converted to Cartesian coordinates and referred to as $P_{rob}$ for further use. The format of $P_{rob}$ is given in Equation \ref{eqn:pose_robot}, and its respective components are shown in Equations \ref{eqn:pose_robot_x},\ref{eqn:pose_robot_y} and \ref{eqn:pose_robot_a}.
\begin{equation}
\label{eqn:pose_robot}
 P_{rob} = \{P_x,P_y,P_a\}
\end{equation}
where,
\begin{equation}
\label{eqn:pose_robot_x}
 P_x = -\sin\theta_{m}-r_{rob}*\cos\theta_y-\sin\theta_{m}*r_{rob}*\sin\theta_y
\end{equation}
\begin{equation}
\label{eqn:pose_robot_y}
 P_y = \sin\theta_{m}*r_{rob}*cos(\theta_y)+\cos\theta_{m}*r_{rob}*\sin\theta_y
\end{equation}
Finally, the orientation of the marker in the environment $\theta_{m}$ specified in the marker position database is added to get the actual angle of the mobile robot with respect to the environment. $\pi$ is added as the camera is looking towards the marker. The calculation of the robot heading is given in Equation \ref{eqn:pose_robot_a}.
\begin{equation}
\label{eqn:pose_robot_a}
 P_a = \theta_{p}+\theta_{y}+(\theta_{m}+\pi)
\end{equation}

\noindent To keep the turret camera pointed to the marker at all times, we use the rotation information from Equation \ref{eqn:marker_pose}. The x-axis rotation component $\theta_x$ is used to calculate the pan angle, and the z-axis rotation component $\theta_z$ is used for calculating the tilt angle. The pan and tilt angles calculated are sent to the turret controller using serial communication over a USB port.
\\
\noindent During the movement of the turret, we observed that the ArUco marker detection loses track of the marker in intermediate frames. Hence, an additional MOSSE Tracker \cite{Bolme2010} is implemented using OpenCV to keep note of the marker in case of intermediate frames being missed by the ArUco marker detection. The turret location is calculated only when the position estimate is available from the ArUco library, and the MOSSE tracker is tracking the marker. The MOSSE tracker also helps compensate for non-identification of the marker by the ArUco library due to camera autofocus issues during the movement of the turret and/or the robot.
\\
\noindent In case of multiple markers being detected in a single frame, only markers available within the dictionary will be considered for calculation of the location. The MOSSE tracker tracks the marker closest to the camera as given by the relative pose. The tracker tracks the current marker unless another marker in the dictionary comes closer to the camera by a certain threshold (20cm). This is not an issue during the mapping stage, as the relationship between two markers is calculated during exploration.
\\
\noindent Localization from the marker tracker module is published as a geometry\_msgs/PoseStamped type topic on ROS. The current pan and tilt angle on the turret received as feedback is also published over ROS for the mapping node for pose estimation.

\item \textbf{Turret Working}
The turret controller is programmed using Arduino IDE. The turret controller awaits commands from the software regarding the target pan and tilt angles. The commands are of the format: $\lt$pan\_target$\gt$,$\lt$tilt\_target$\gt;$. The controller also reports the current state of the pan and tilt servo motors to the software using a similar format ($\lt$pan\_state$\gt$,$\lt$tilt\_state$\gt$;). We first check if the target angles are beyond the physical limits of the setup. The pan angle is restricted between (10°, 350°) and the tilt angle is limited to (-70°, 90°). The corresponding values mapped to these angles are pan (100, 3980) and tilt (1248, 3072). If the target angles are beyond these constraints, the maximum permissible angle is input to the tilt servo. For the pan servo, target angles beyond the limit cause the servo to rotate in the opposite direction so as to catch up with the marker. For e.g., with current angle = 350° and input as 370°, the command is converted to 10°, and the turret rotates 340° in the opposite direction.\\
\noindent We implement a PID controller to ensure smooth transitions between the current state and the target state of the turret without any oscillations. Input to the PID controller is the current state of the servo motor and the target angle received over serial communication. The difference between these two states is considered as the error factor in the equation governing the PID control variables (Equation \ref{eqn:PID}). The two servo motors are controlled by different PID equations.
\begin{equation}
\label{eqn:PID}
u(t)=K_pe(t)+K_i\int e(t)d(t) + K_d \frac{de}{dt}
\end{equation}
where,\\
$u(t)$ is the control signal to be sent to the servos,\\
$e(t)$ is the error between the current and target state,\\
$K_p,K_i, K_d$ are constants governing the proportional, integral, and differential components of the equation. Through experimentation, we found that $K_p=0.7,K_i=0, K_d=0$ are ideal values for our application.
We have a provision to reset the turret on the key press. The turret can also switch to specified pan angles immediately for transitioning between markers during navigation. These manual overrides are useful during the mapping stage and for transitioning between markers during navigation.

\item \textbf{Marker to Marker Transforms}
\label{sec:marker_trans}
During exploration of the environment, the map of the environment is generated in 2D. Hence, the pose of the marker relative to the robot is converted to 2D cartesian coordinates using the Equations \ref{eqn:polar_pose}, \ref{eqn:polar_r} and \ref{eqn:polar_theta}. For associating markers detected in the environment during the exploration, we use the following method. The first marker detected by the system $M_{0}$ is considered to be at the origin of the environment facing in the negative y-direction $M_0=(0,0,180^\circ)$. Subsequent markers are placed within the experimentally determined accurate range of the first marker. The robot is operated manually and is taken around one marker for exploration. The turret camera and the software together keep track of the first marker and perform mapping based on the localization information derived from tracking the marker (from Equation \ref{eqn:pose_robot}). For transition between markers, the robot is brought to the transition zone, i.e., the area of the environment that is within reach of both the markers under consideration. Here, the robot is stopped, and a manual signal triggers the turret to reset and start tracking the new marker. The location of the robot in relation to the previous marker is saved when the robot is stopped. Transform between the robot's coordinates to the first marker's coordinates is represented by Equation \ref{eqn:trans_r0}.
\begin{equation}
\label{eqn:trans_r0}
T_{R0} =     \begin{bmatrix}
        \cos{P_a} & -\sin{P_a} & P_x\\
        \sin{P_a} & \cos{P_a} & P_y\\
        0 & 0 & 1\\
    \end{bmatrix}
\end{equation}
where $P_x, P_y$ and $P_a$ are from Equations \ref{eqn:pose_robot_x}, \ref{eqn:pose_robot_y} and \ref{eqn:pose_robot_a}.
This is used to calculate the relative position of the new marker for further localization and hence, mapping. The transform between the robot's coordinates to the new marker $M_1$ is given in Equation \ref{eqn:trans_r1}.
\begin{equation}
\label{eqn:trans_r1}
T_{R1} =     \begin{bmatrix}
        \cos{\theta_{M1}} & -\sin{\theta_{M1}} & r_{rob}\cos\theta_{p}\\
        \sin{\theta_{M1}} & \cos{\theta_{M1}} & r_{rob}\sin\theta_{p}\\
        0 & 0 & 1\\
    \end{bmatrix}
\end{equation}
where $\theta_{M1}=\theta_{p}+\theta_y+\pi$ with $\theta_{p}$ being the pan angle and $\theta_y$ is from the marker pose in Equation \ref{eqn:marker_pose}.
The composite transform to yield the transform between markers $M_1$ and $M_2$ is given in Equation \ref{eqn:trans_12}.
\\
$T_{12} =     $
\begin{equation}
\label{eqn:trans_12}
\begin{bmatrix}
        \cos{P_{aM1}} & -\sin{P_{aM1}} & r_{rob}\cos{(\theta_{p}+P_a)}+P_x\\
        \sin{P_{aM1}} & \cos{P_{aM1}} & r_{rob}\sin{(\theta_{p}+P_a)}+P_y\\
        0 & 0 & 1\\
\end{bmatrix}
\end{equation}
where, $P_{aM1}=P_a+\theta_{M1}$.
The robot is made to explore around marker $M_{1}$ and reach the transition zone to another marker (say, $M_{2}$) and starts tracking that marker. This process is repeated until the accessible sections of the environment are mapped.
\end{itemize}

\subsubsection{Environment Mapping}
\label{sec:env_mapping}
\noindent 2D mapping of the environment is performed to identify objects that may be missing in the floor plan of the building. Architectural drawings of the buildings fail to capture necessary features, e.g., furniture \cite{Cheraghi2017}. Here, we use the mapping robot configuration as shown in Figure \ref{fig:mapping_robot}. The robot is manually operated to cover as much of the environment as possible in one go.

\noindent A 2D map of the environment is created for marker placement and path planning. The environment is represented in the form of occupancy grid maps. Occupancy grid-type maps are represented as a matrix where a value between 0 to 100 represents the probability in percentage of the cell being an obstacle. Unknown regions of the environment are represented by -1. In ROS, occupancy grid maps are represented as message type nav\_msgs.msg/OccupancyGrid\footnote{\url{http://docs.ros.org/en/kinetic/api/nav_msgs/html/msg/OccupancyGrid.html}}. In our approach, we use the marker-based localization for mapping in an unknown environment with visual markers. The markers are placed in a way that they are within tracking range of each other (4m). Since we already have the marker tracker implementation in place estimating localization, we can utilize that along with the range information from the laser scanner to build the occupancy grid map. The laser scanner used is a Hokuyo URG-04LX-UG01. Data from this sensor is accessed using the ROS Package urg\_node \footnote{\url{http://wiki.ros.org/urg_node}} on the topic /scan. It is of the type ROSMSG: sensor\_msgs/LaserScan. The readings from the sensor represent the distance between the laser scanner and the nearest obstacle.
\\
\noindent We compare our approach with a commonly used state-of-the-art mobile robot-based SLAM for map generation. Hector SLAM \cite{Kohlbrecher2011} works on the principle of scan matching in combination with a 6DOF motion estimation using inertial sensing. High scan rates of modern laser scanners can be utilized for registering patterns from consecutive scans, thus yielding estimates for 2D position and orientation. Data from the accelerometer and gyroscopes are used as input to an Extended Kalman Filter (EKF) for estimation of the 6D pose of the platform. This method is tested to work in real-time with decent computing resources. We used the ROS implementation of Hector SLAM from its official ROS Package\footnote{\url{http://wiki.ros.org/hector_slam}}.
\\
\noindent Modern implementations of LIDAR-based mapping like \cite{zhang2014loam} and \cite{liu2021balm} are not included in this study. Zhang et al.'s approach \cite{zhang2014loam} relies on a LIDAR mounted on a tilt mechanism to help capture data in its vertical field-of-view. Liu et al. \cite{liu2021balm} extend the previous approach with Bundle Adjustment to compensate for the increased amount of data generated by the 3D LIDAR used. 3D mapping is not of primary concern for us since we are focused on navigation alone, and that can be achieved with 2D maps of the environment. The sensor setups for both these approaches are cumbersome or expensive. In addition, these approaches still suffer from the long corridor problem, where the absence of features within the range of the LIDAR cripples the pose estimation.
\\
\noindent We use the laser scanner data and the localization data to construct two maps of our surroundings. The ``Local Map'' stores the ray-traced pattern of the laser scanner's output. It considers the laser scanner as the origin and conveys the live feed of the data in a visual sense. The ``Global Map'', takes into account the transformation between the laser scanner frame and the localization information of the robot and appends the new data to construct the complete map of the environment.

\begin{itemize}
\item \textbf{Processing Raw Laser Scanner Data} 
The laser scanner data contains noise in the form of NaN and infinity values. Infinity values are eliminated by thresholding the data to within 3.5 meters. NaNs instances are assigned values based on neighboring range values using linear interpolation. The laser scanner's range data contains of range readings starting from the right side of the laser, covering the entire FOV of the scanner in an anti-clockwise direction. We derive the polar coordinates of the laser scanner readings and convert them to cartesian coordinates to perform raytracing for map building (Equations \ref{eqn:polar_pose}, \ref{eqn:polar_r} and \ref{eqn:polar_theta}).

\item \textbf{Plotting the Live Map}
The live map stores the incoming laser scanner data in the form of range-bearing data $z_{t}^{i}=(r_{t}^{i},\theta_{t}^{i})^{T}$. Given the radial coordinates of the range-bearing information, we can draw lines to that point from the origin (laser scanner is the origin in the live map) to that point using the Equation \ref{eqn:laser}.
\begin{equation}
\label{eqn:laser}
    (r_{t}^{i},\theta_{t}^{i})^{T}=\frac{\sqrt{(m_{j,x}-x)^2+(m_{j,y}-y)^2}}{atan2(m_{j,y}-y,m_{j,x}-x)-\theta}+Q_{t}
\end{equation}
The occupancy grid map is a discrete representation of the environment, and the scale of the occupancy grid map is set to 20 pixels/meter. The cartesian coordinates are converted to discrete coordinates before ray tracing. The dimensions of the occupancy grid map are initialized to 1000x1000 cells, with all cells set to -1 (unknown region). Range readings that are $\lt$3.5 meters are assigned a value of 100 (obstacle), whereas we assign the value 50 (possible obstacle) to all 8 connected cells next to that cell. Cells between the origin and the range readings are assigned a value 0 to indicate free space.

\item \textbf{Constructing the Global Map}
The global map is a collection of local maps stitched together using localization information. For each cell within range of the laser scanner at the pose of the robot, we take the average reading of all its historic values. Since the map update is probabilistic, any sharp changes to the values are smoothed out. A matrix maintaining the count of all the readings in the current session is also maintained for the calculation of the occupancy grid cell probability. The update step of a cell is specified in Equation \ref{eqn:map_update}.
\\
$p(m_i\|z_{1:t},x_{1:t}) =$
\begin{equation}
\label{eqn:map_update}
\frac{p(m_i\|z_{t},x_{t})}{1-p(m_i\|z_{t},x_{t})}*\frac{p(m_i\|z_{t-1},x_{t-1})}{1-p(m_i\|z_{t-1},x_{t-1})}*\frac{1-p(m_i)}{p(m_i)}
\end{equation}
where,\\
\noindent $m$ is the matrix representing the current occupancy probability of the environment,\\
\noindent $z$ is the sensor observation model of the laser scanner, and \\
\noindent $x$ represents the localization information of the robot corresponding to the global map.
The method used is the one proposed by Moravec et al. \cite{moravec1985high} for mapping with known poses.
\end{itemize}
\subsection{Map Preparation and Marker Placement}
Map preparation is an intermediate step between the mapping phase and then navigation phase that helps in reducing the size of the map. Marker placement reduces the number of markers required to cover the environment to provide localization to the BVI user.

\subsubsection{Map Decomposition}
\begin{figure}[!t]
\centering
\includegraphics[width=\columnwidth]{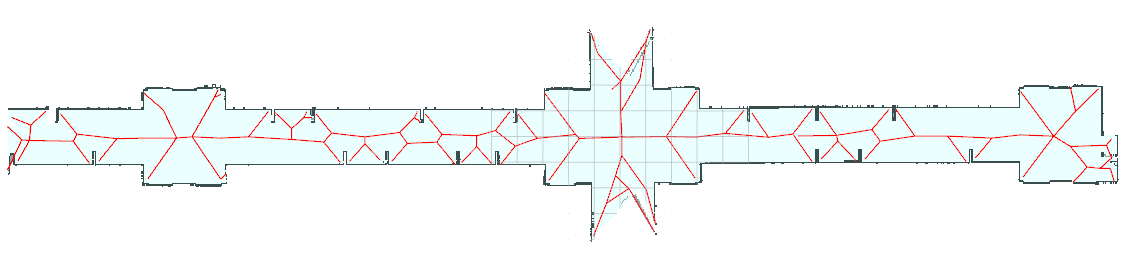}
\caption{Voronoi partitioning applied on corridor environment}
\label{fig:voronoi_part}
\end{figure}

\begin{figure}[!t]
\centering
\includegraphics[width=\columnwidth]{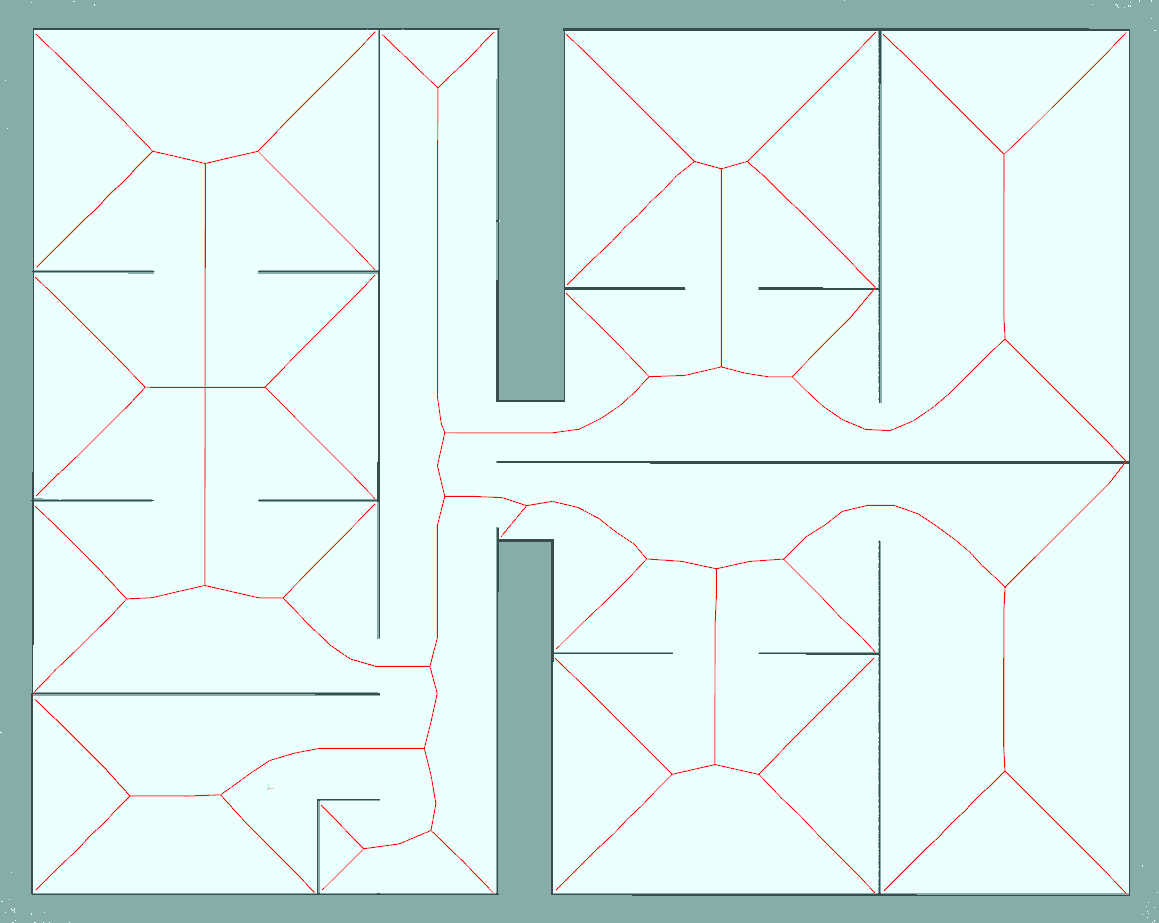}
\caption{Voronoi partitioning applied on lab environment}
\label{fig:dt_voronoi_part}
\end{figure}

\noindent Map decomposition is used in robotics to build compact representations of the environment as well as to derive information on open spaces in the environment. VA systems using maps for navigation rely on a variety of methods for map representation and path planning. Examples range from XML-based \cite{Ivanov2017} to complete CAD models \cite{Li2019} for known static maps and occupancy grid-based to point cloud-based representation \cite{DelaPuente2019} for systems with live mapping. We use Voronoi-based partitioning to identify the boundaries of the Voronoi regions in the environment. The Voronoi decomposition approach is inspired by Wang et al. \cite{Wang2013}. They convert the occupancy grid map to a clearance map by Euclidean Distance Transform (EDT). Cell boundaries are generated by thinning the clearance map. These boundaries represent the parts of the environment that are at maximum distance from the obstacles in the vicinity. Hence, these points can be used for navigation in the environment. Voronoi decomposition results of the generated maps are shown in Figures \ref{fig:voronoi_part} and \ref{fig:dt_voronoi_part}.

\begin{itemize}
\item\textbf{Reduced Marker Placement}
\label{sec:reduced_markers}
\noindent We propose a novel marker placement methodology for ensuring minimal defacing of the environment. Given an occupancy map of the environment, our algorithm proposes coordinates for the placement of the fiducial markers. We use a novel formulation of the  art gallery problem for solving optimal area coverage \cite{MikaelPalsson2008}. This method is divided into three sections: (1) Rectangular Decomposition of the environment, (2) Generation of Candidates for Marker Placement, and (3) Selective Marker Removal.

\item \textbf{Rectangular Decomposition}
\noindent The Fiducial markers used in this work have all-around visibility, hence we segment the environment into rectangles. The centroids of the rectangular sections generated are possible candidates for the placement of a marker. Rectangular decomposition also aligns with our assumption of an orthogonal indoor environment.
\\
Inputs to the algorithm include the occupancy grid map of the environment. The grid representing the environment is a 2D matrix of size 1920x1080 with a resolution of 20 pixels/meter. A rectangle is defined as R=\{x,y,w,h\} where x and y are the coordinates of the centroid of the rectangle, and w and h represent the width and height of the rectangle, respectively.
\\
The process of rectangular decomposition is mentioned in Algorithm \ref{alg:rect}. Firstly, we preprocess the occupancy grid map (Line \ref{alg:rect:preproc}). The occupancy grid map undergoes thresholding to create a binary map. Applying the skeletize operation on the binary map obtained yields a map with single-pixel boundaries for easy edge detection. Next, the processed map is scanned from left to right (Line \ref{alg:rect:scan}), identifying each vertical edge in the orthogonal map. As soon as an edge is detected, its corresponding closing edge is scanned in the sub-image obtained by cropping the complete map to the size of the edge (Line \ref{alg:rect:scansub}). On detection of the closing edge or the end of the map (Line \ref{alg:rect:close}), a rectangle based on the starting and ending edge is saved (Line \ref{alg:rect:save}). This process is repeated till all vertical edges are identified and closed to form rectangles.
\\
After all rectangles have been identified, pruning is done to remove rectangles identified in the non-traversable region of the environment (Line \ref{alg:rect:prune}). Also, the rectangle boundaries are trimmed to ensure they lie within the traversable region of the environment (Line \ref{alg:rect:prune:trim}). Refer to section \ref{sec:plac_res} and Figures \ref{fig:rectangulate} and \ref{fig:dt_lab_rectangulation} for results.

\begin{algorithm}
\caption{Rectangular Decomposition of Occupancy Grid.}\label{alg:rect}
\begin{algorithmic}[1]
\renewcommand{\algorithmicrequire}{\textbf{Input: }}
\renewcommand{\algorithmicensure}{\textbf{Output: }}
\Require Occupancy grid map of Orthogonal environment $G_{m}$.
\Ensure Rectangles representing contiguous sections of map $R_{k}$.
\State Erode the map boundaries to obtain single pixel width boundaries.\label{alg:rect:preproc}
\For {width of $G_{m}$}\label{alg:rect:scan}
    \If {edge detected}
        \State Initialize New Rectangle $R_{i}$
        \State {Store the edge as an edge of the rectangle $R_{i}=\{x,y,0,h\}$}
        \For {Width of Subsection}\label{alg:rect:scansub}
            \If {closing edge detected}\label{alg:rect:close}
                \State {Store closing edge of Rectangle  $R_{i} \Leftarrow \{x,y,w,h\}$}\label{alg:rect:save}
            \EndIf
            \If {Map End reached}
                \State {Store closing edge of Rectangle  $R_{i} \Leftarrow \{x,y,w,h\}$}
            \EndIf
        \EndFor
    \EndIf
    \State {Store Rectangle Generated $R_{k} \Leftarrow R_{i}$}
\EndFor
\For {All Rectangles $R_{k}$}\label{alg:rect:prune}
    \If {$R_{i}$ lies in traversable region}
        \State Trim $R_{i}$ to traversable region\label{alg:rect:prune:trim}
    \Else
        \State Remove $R_{i}$ 
    \EndIf
\EndFor
\State return $R_{k}$\label{alg:rect:return}
\end{algorithmic}
\end{algorithm}

\item \textbf{Candidate Generation}
\noindent The centroid of the rectangular sections are considered candidates (refer Algorithm \ref{alg:candidates}) for marker placement such that one marker with circular range can cover the entire rectangle. The range of the visual marker is an integer representing the radius of the region around the marker where the tracker can give accurate localization. The range of the marker is determined empirically in Section \ref{section:marker_range}. The range is multiplied by the resolution and is considered an integer number representing the number of pixels for the range in meters. Contiguous areas in indoor spaces may be bigger than the range of any visual marker. In such a case, the placement of the marker at the centroid of the rectangle will be rendered futile. To address this, rectangular sections obtained from Algorithm \ref{alg:rect} Line \ref{alg:rect:return} are checked if their dimensions exceed the range of the marker (Algorithm \ref{alg:candidates} Lines \ref{alg:candidates:checkw} and \ref{alg:candidates:checkb}). If this happens, the rectangle is segmented along that dimension to yield individual segments smaller than the range of the marker (Lines \ref{alg:candidates:segw} and \ref {alg:candidates:segb}). If it is observed that the dimension is less than half of the given range, the vertices of the rectangles are also added to the list of candidates for marker placement (Line \ref{alg:candidates:corners}) along with the centroid of the rectangle (Line \ref{alg:candidates:centroids}). This is to ensure that the marker placed in the rectangle has improved coverage of the environment. The marker placement candidates are represented as 2D coordinates in the occupancy grid map M=\{x,y\}.

\begin{algorithm}
\caption{Marker Placement Candidate Generation}
\label{alg:candidates}
\begin{algorithmic}[1]
\renewcommand{\algorithmicrequire}{\textbf{Input: }}
\renewcommand{\algorithmicensure}{\textbf{Output: }}
\Require Rectangles representing contiguous sections of map $R_{k}$, range of visual marker $r$.
\Ensure Comprehensive set of locations for placing markers $M_{c}$.
\State {Initialized Segmented Set $R_{s}$}
\For {All Rectangles $R_{k}$}
    \If {$r \lt width(R_{i})$}\label{alg:candidates:checkw}
        \State {$R_{s} \Leftarrow $ Segmented rectangle over width}\label{alg:candidates:segw}
    \EndIf
    \If {$r \lt height(R_{i})$}\label{alg:candidates:checkb}
        \State {$R_{s} \Leftarrow $ Segmented rectangle over breadth}\label{alg:candidates:segb}
    \EndIf
\EndFor
\State $R_{k} \Leftarrow R_{s}$
\State Initialize List of Candidates $M_{c}$;
\For {All Rectangles $R_{k}$}
\If {$width(R_{i})<r and height(R_{i})<r$}\label{alg:candidates:corners}
\State {$M_{c} \Leftarrow $ corners of $R_{i}$}
\EndIf
\State {$M_{c} \Leftarrow $ centroid of $R_{i}$}\label{alg:candidates:centroids}
\EndFor
\State return $M_{c}$
\end{algorithmic}
\end{algorithm}

\item \textbf{Selective Marker Removal}
\label{sec:marker_opt}
\noindent Candidate generation for marker placement in the previous section generates multiple markers in contiguous sections of the environment due to the constraints on the range and rectangular decomposition of the map. In order to reduce the number of markers required while maintaining complete coverage, we follow an iterative removal of the markers (Algorithm \ref{alg:optimize}). Given the list of possible marker placement locations, we sort them by their distance from the nearest obstacle (Line \ref{alg:optimize:sort}). This works on the logic that the marker furthest from the obstacle can make better use of its 360° capability and cover a larger area in the map than being closer to an obstacle. This is especially useful considering that we need to reduce the number of markers. The above logic is illustrated in Figure \ref{fig:marker_sort} where we see that the marker farther away from the wall M2 covers more cells (2380) than M1 (2360).
\begin{figure}[!t]
\centering
\includegraphics[width=\columnwidth]{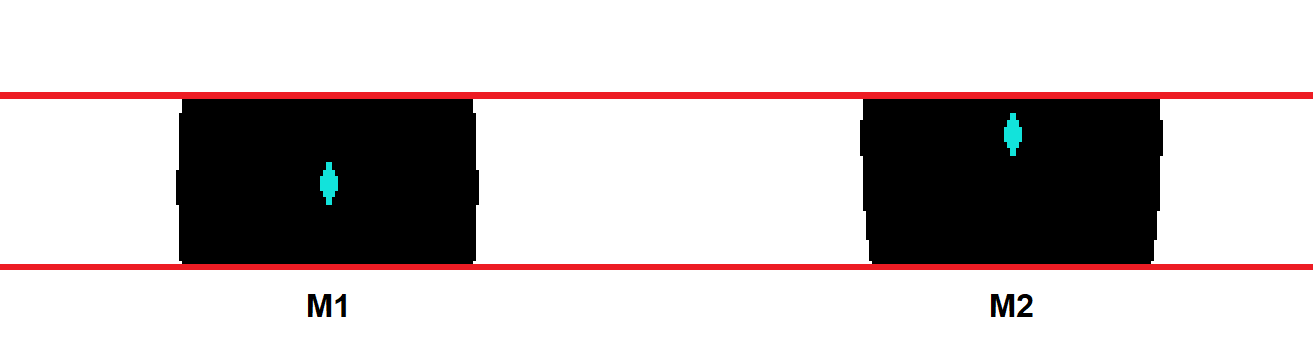}
\caption{In a long corridor, marker M1 is close to edge and covers 2360 cells and M2 away from edge covers 2380 cells}
\label{fig:marker_sort}
\end{figure}
Next, we generate the ray-traced patterns to determine the area of the map covered by each marker. Further, we remove each marker (Line \ref{alg:optimize:remove}) and recalculate (Line \ref{alg:optimize:recalc}) the coverage and overlap for the remaining markers. The combination that has the maximum amount of overlap is retained and sent for the next iteration. This is continued till the removal of a marker from the list results in loss of coverage (Line \ref{alg:optimize:reiter}).

\begin{algorithm}
\caption{Reduced Marker Placement}
\label{alg:optimize}
\begin{algorithmic}[1]
\renewcommand{\algorithmicrequire}{\textbf{Input: }}
\renewcommand{\algorithmicensure}{\textbf{Output: }}
\Require Comprehensive set of locations For placing markers $M_{c}$.
\Ensure Locations For reduced placement of markers $M_{o}$.
\State Sort Marker candidates $M_{c}$ based on distance to nearest obstacle; \label{alg:optimize:sort}
\State Generate Ray-traced maps of the area covered by each marker in $M_{c}$ as $RT_{c}$;
\State $M_{t} \Leftarrow M_{c}$;

\For {All Marker Positions $M_{t}$}
\State $M_{t} \Leftarrow M_{c} \cap M_{i}$ \label{alg:optimize:remove}
\State {Evaluate coverage and overlap with $M_{t}$} \label{alg:optimize:recalc}
\State {Find Index with Maximum Coverage and Minimum Overlap}
\EndFor
\If {$C_{i}=1$}\label{alg:optimize:reiter}
\State {$M_{c} \Leftarrow M_{t}$}
\State {Repeat Step 3}
\Else
\State return {Reduced Marker Positions $M_{o}$}
\EndIf

\end{algorithmic}
\end{algorithm}
\end{itemize}
\subsubsection{Path Points to Marker Association}
\label{sec:path_marker_assoc}
\noindent Localization information obtained using the marker tracker can be used to track entities in the indoor environment and provide navigation assistance. For our navigation stack, we associate the path points determined by the Voronoi partition boundaries with the marker placed nearest to it. Occupancy grid points lying on the Voronoi partition are assigned a marker id based on the ray-traced maps of the cells covered by the reduced set of markers obtained in Section \ref{sec:marker_opt}. In case of a conflict of marker assignment - for points lying in the overlapping region of multiple markers - the marker with the least Euclidean distance is assigned to that point on the path.

\subsection{Navigation Phase}
\begin{figure}[!t]
\centering
\includegraphics[width=\columnwidth]{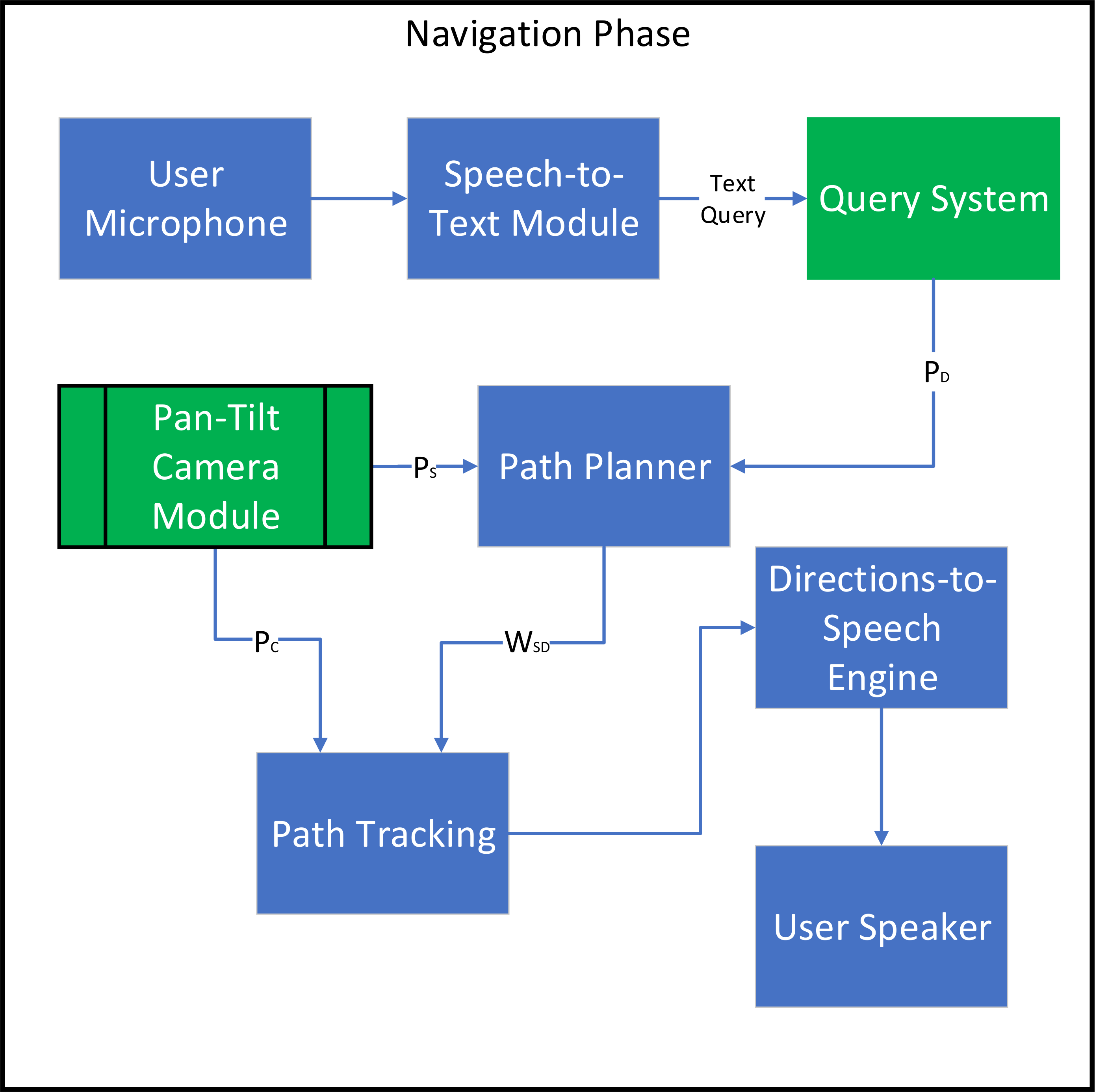}
\caption{High-level block diagram of the navigation system. $P_{C}$ conveys the pose of the user derived from the PT camera Module. $P_{S}$ and $P_{D}$ represent the source and destination pose respectively. $W_{SD}$ indicates the waypoints generated by the path planner from the source S to destination D}
\label{fig:Navigation_BD}
\end{figure}

\noindent 
The BVI user interacts with our system using audio input/output (Figure \ref{fig:Navigation_BD}).
\begin{itemize}
    \item The user microphone captures audio commands that convey the place of interest for navigation.
    \item The Query System looks up the destination in the Map database based on the features labeled in the environment.
    \item The current location of the user given by the marker-based localization system acts as the source, and the location returned by the Query System is set as the destination and passed to the Path Planner.
    The Navigation Module acts up on the path planned and current user location from the marker-based localization and triggers appropriate navigation commands to the BVI user.
\end{itemize}

\subsubsection{Navigation Mode Turret Operation}
\label{sec:navig}
\noindent For navigation, marker locations are known apriori. Hence, transformations between markers are known. To ensure smooth autonomous navigation in the environment, it is important to implement an autonomous tracker transition between one marker to another. Since the closest marker yields the best localization data, the marker transition is designed on the basis of Euclidean distance from the location of the robot to all markers in the marker position database. The robot starts with any marker in its FOV (say, $M_{init}$), and it gets an initial location estimate $P_{rob}$. Given the pose of all markers in the environment, we determine the nearest marker based on Euclidean distance (Equation \ref{eqn:nearest} and \ref{eqn:dist_rm}).
\begin{equation}
\label{eqn:nearest}
M_{near}={arg\,min}_j dist(P_{rob},M_j)
\end{equation}
where,
\begin{equation}
\label{eqn:dist_rm}
dist(P_{rob},M)=\sqrt[]{(P_x-M_x)^2+(P_y-M_y)^2}
\end{equation}
The turret points to $M_{near}$ based on the pan angle derived from the current heading of the robot $P_a$ and the coordinates of the marker and the robot (Equation \ref{eqn:r_theta}).
\begin{equation}
\label{eqn:r_theta}
\theta_p=P_a+atan((P_y-M_y)/(P_x-M_x))
\end{equation}
Once the robot starts moving towards a specified destination, this calculation runs continuously to ensure that the turret always points to the nearest marker.

\subsubsection{Path Planner}
\label{sec:path_planner}
\begin{figure*}[!t]
\centering
\includegraphics[width=\textwidth]{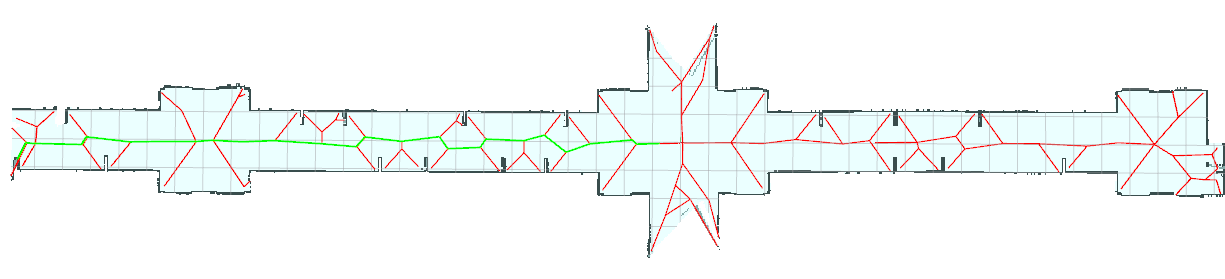}
\caption{Path planning example on the Voronoi partition lines}
\label{fig:voronoi_path}
\end{figure*}

\noindent Voronoi region boundaries represent points in the environment used for navigation, a path planner considering this set of points is used. Any source or destination not on the Voronoi region boundaries is mapped to the nearest point. Path Planning using the given Voronoi boundary graph points is achieved by a specialized version of Dijkstra’s algorithm as specified in \cite{Wang2013}.
\\
\noindent Path planning on the Voronoi boundaries for the corridor map is shown in Figure \ref{fig:voronoi_path}. The green line in the image denotes the path from the source at the center of the environment and the destination at the left corner.

\subsubsection{Robot Navigation}

\noindent The output of the path planner in section \ref{sec:path_planner} is a set of waypoints that are along the Voronoi boundaries. Navigation to individual waypoints is achieved through basic motion of the robot comprising two rotations and one translation. Given source and destination points $ S=(x,y,\theta) $ and $ D=(x', y', \theta') $, the calculations for traversal distance, initial and final rotation are represented by Equations \ref{eqn:nav_dist}, \ref{eqn:nav_theta1} and \ref{eqn:nav_theta2} respectively.
\begin{equation}
\label{eqn:nav_dist}
\delta_{trans}=\sqrt{(x^{'}-x)^{2}+(y^{'}-y)^{2}}
\end{equation}
\begin{equation}
\label{eqn:nav_theta1}
 \delta_{\theta1}=atan((y'-y)/(x'-x))-\theta
\end{equation}
\begin{equation}
\label{eqn:nav_theta2}
 \delta_{\theta2}=\theta'-\theta-\delta_{\theta1}
\end{equation}
With $\delta_{\theta1}$ representing the first rotation to get the robot on the trajectory, $\delta_{trans}$ representing the distance to be traveled on the trajectory, and $\delta_{\theta2}$ being the angle of rotation for the robot to achieve the desired heading $\theta'$ specified by the destination point. Navigating to each waypoint in order ensure that the mobile robot reaches the destination.
\\
\noindent Pure pursuit controller implementation is from Coulter et al. \cite{coulter1992implementation}. We obtain the current location of the agent from the marker tracker pose estimate. The path planner provides the waypoints between the current location of the agent and the goal location. Next, we set a lookahead distance of 1 meter and find the path point closest to the agent. Velocity commands are generated and sent to the robot to steer toward the nearest path point. The goal point -  considered to be the next waypoint to be reached is determined and converted to the robot's frame of reference, and curvature between the points is obtained. Velocity commands to steer the vehicle are issued based on this curvature. This cycle is repeated for each navigation cycle corresponding to each waypoint.

\subsubsection{BVI Navigation}

\noindent Robot navigation described in the previous section was conducted to test the localization and tracking performance while a robot travels a path defined by the path planner. This section describes the implementation of the VA system on a BVI individual and its working. Continuing with the output of the path planner, we have a list of waypoints representing the path from the current location of the user to the input destination. Using the Equations \ref{eqn:nav_dist}, \ref{eqn:nav_theta1} and \ref{eqn:nav_theta2}, we obtain the angles and the distance between consecutive waypoints. The user will be given directions in a similar fashion to the robot navigation. First, the user will rotate $\delta_{\theta1}$ degrees to point in the direction of the next waypoint. For this, the audio output prompts the user to rotate slowly in the right or left direction till the correct heading is achieved. Next, for covering the linear distance $\delta_{trans}$ to the destination, the user is prompted to go straight till the waypoint is reached. To achieve the second rotation $\delta_{\theta2}$, we follow the same procedure as for $\delta_{\theta1}$. This cycle is repeated for all waypoints till the destination is reached.

\section{Experiments}
\noindent Experiments were conducted to verify the accuracy and applicability of the novel localization technique for indoor environments. This section contains experiments performed to design the marker tracker system and presents the evaluation of individual components, namely mapping and navigation.
\label{sec:experiments}
\subsection{Hardware Configuration}
The hardware configuration for the experiments conducted uses a mobile robot platform for locomotion of the range sensor to map the environment. For implementing the visual marker tracking, we fixed the PT turret atop the mobile robot. In addition to that, we printed out the markers as per the custom dictionary and fixed them on cardboard boxes suspended from the ceiling.
\subsubsection{Mobile Robot Platform}
\noindent The mobile robot platform used for the experiments is the Firebird VI by NEX Robotics\footnote{\url{http://www.nex-robotics.com/products/fire-bird-vi-robots/fire-bird-vi-robotic-research-platform.html}}. It is available in different drive variants; we used the one with 4-wheel differential drive. The robot is equipped with IR and SONAR sensors for long and short-range sensing, respectively. Computing is handled by an Intel NUC-based mini-PC with a laptop-grade Core i5-7260U processor and 8 GB of RAM running Ubuntu 16.04 and ROS Kinetic. We installed a Hokuyo URG-04LX-UG01 Laser Scanner on the robot. This offers a 240° FOV around the robot and has a range of 5 meters. We clipped the range to 3.5 meters to reduce the error  over large distances by the laser range finder. The robot is controlled over USB serial communication to the built-in microcontroller using its ROS Node. Wheel odometry and range sensor information are also available over the respective ROS Nodes.
\subsubsection{Pan-Tilt Turret}
\noindent The PT turret on which the camera is mounted is a WidowX MX-28 PT turret from Trossen Robotics. It consists of two Dynamixel MX-28 servo motors connected such that one servo changes the pan angle and another serves the tilt function. The two servos are controlled by an Arbotix-M Robocontroller board which takes input from the PC via USB for the PT angles and supplies power to the servos. An additional 12V Li-ion battery is provided for the functioning of the turret.

\subsection{Marker Range Testing}
\label{section:marker_range}
\begin{figure}[!t]
\centering
\includegraphics[width=\columnwidth]{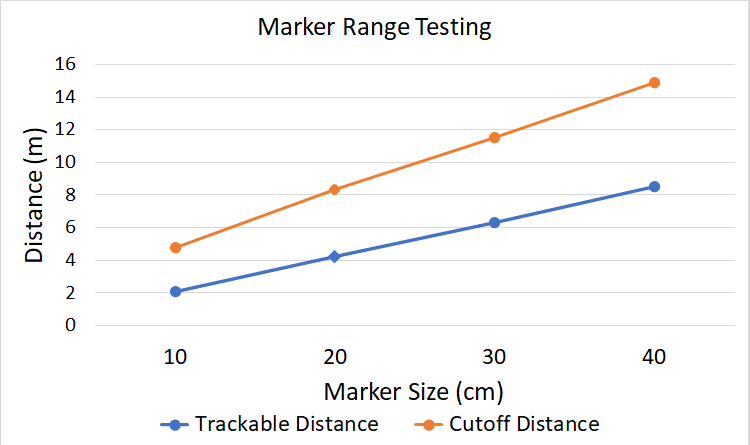}
\caption{Marker range testing graph}
\label{fig:range_graph}
\end{figure}

\noindent Marker range was tested using different-sized markers across various camera resolutions to obtain ranges for each combination. The input image resolution was fixed at 1920x1080 with the camera's built-in auto focus being used. Marker sizes ranging from 10 cm to 40 cm were tested in the same lighting condition and placed at a constant height of 2.5 meters from the floor level to simulate the marker hanging off the roof of the indoor environment. The range testing results give us two metrics, tracking distance, and cutoff distance. Tracking distance is the maximum distance at which the given camera can segment the marker from the environment for the ArUco pipeline to detect the marker with pose estimate at 30 fps. Cutoff distance is the distance from the marker at which the camera is only able to detect the marker without its pose estimate. The test results are tabulated in Table \ref{tab:marker_range}.
\\
\begin{table}
\begin{center}
\caption{Marker range testing results.}
\label{tab:marker_range}
\begin{tabular}{| c | c | c |}
\hline
\textbf{Marker} & \textbf{Tracking} & \textbf{Cutoff}\\
\textbf{Size} & \textbf{Distance} & \textbf{Distance}\\
(cm)&(meters)&(meters)\\
\hline
10& 2.1 & 4.8\\
\hline
20& 4.25 & 8.35\\
\hline
30& 6.3 & 11.5\\
\hline
40& 8.5 & 14.9\\
\hline
\end{tabular}
\end{center}
\end{table}
\\
\noindent As seen from Table \ref{tab:marker_range} and Figure \ref{fig:range_graph}, the results vary linearly based on the physical size of the marker while keeping the resolution of the image constant. However, it must be noted that bigger markers result in more defacing of the environment. Since we are using cuboid-shaped markers for a 360° FOV, the marker size needs to be kept in check.
\subsection{Test Environment}
\begin{figure}[!t]
\centering
\includegraphics[width=\columnwidth]{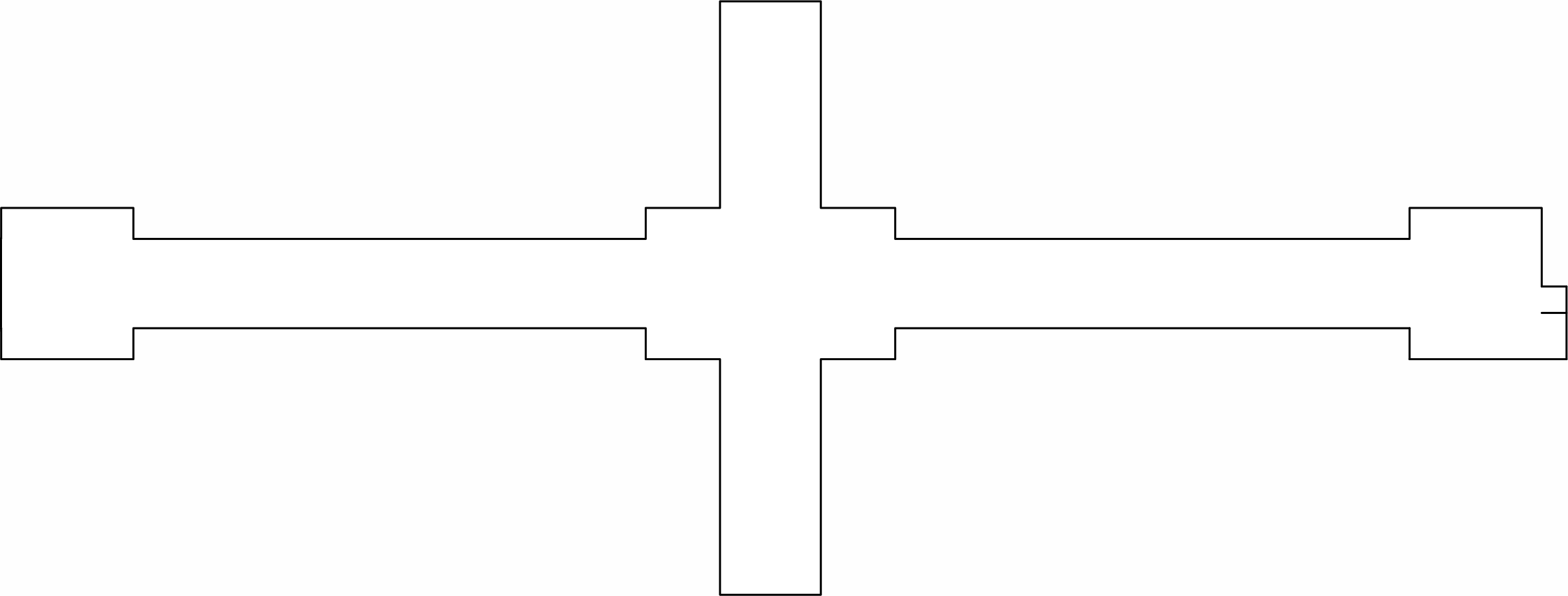}
\caption{Ground truth map of corridor environment of size 1700 sq. ft}
\label{fig:model_map}
\end{figure}

\begin{figure}[!t]
\centering
\includegraphics[width=\columnwidth]{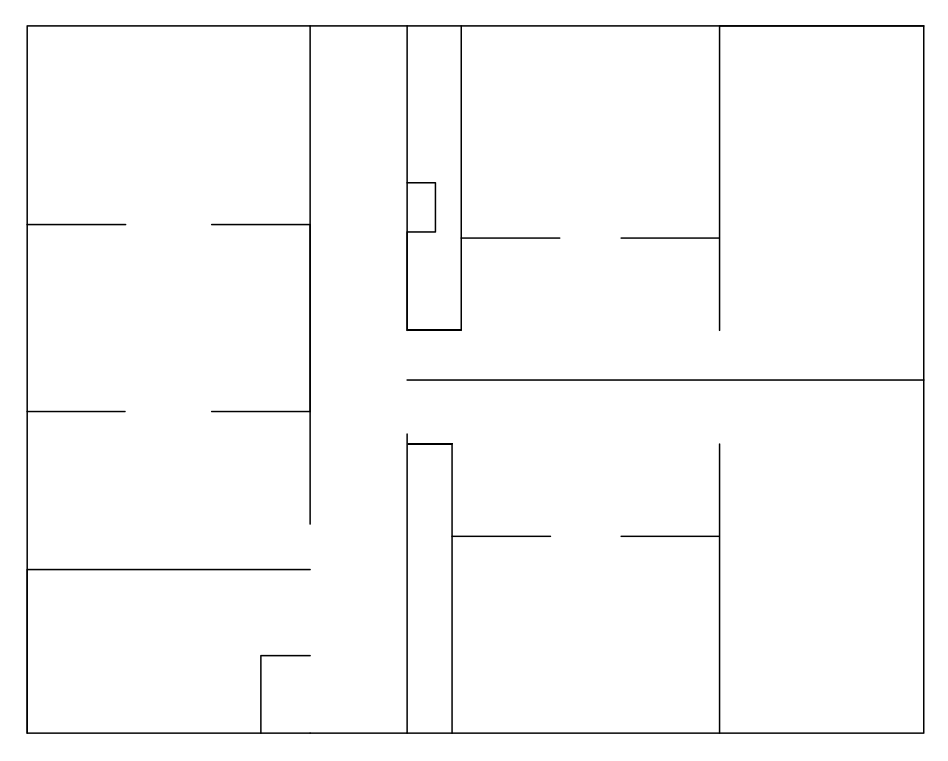}
\caption{Ground truth map of lab environment of size 1800 sq. ft}
\label{fig:dt_map}
\end{figure}

\noindent Real-life mapping and navigation were performed on sections of our university. The first environment consists of a central corridor with one intersection and recesses for door openings. The ground truth map of this environment is shown in Figure \ref{fig:model_map}. The mapped area covers 1700 sq. ft. The second map is a lab with cubicle spaces for seating (Figure \ref{fig:dt_map}). This covers an area of 1800 sq. ft.

\subsection{Mapping Results}
\label{sec:map_res}
\begin{figure}[!t]
\centering
\includegraphics[width=\columnwidth]{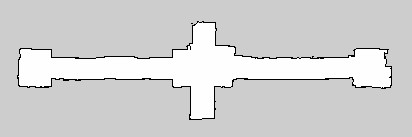}
\caption{Mapping using Hector SLAM}
\label{fig:hector_map}
\end{figure}

\begin{figure}[!t]
\centering
\includegraphics[width=\columnwidth]{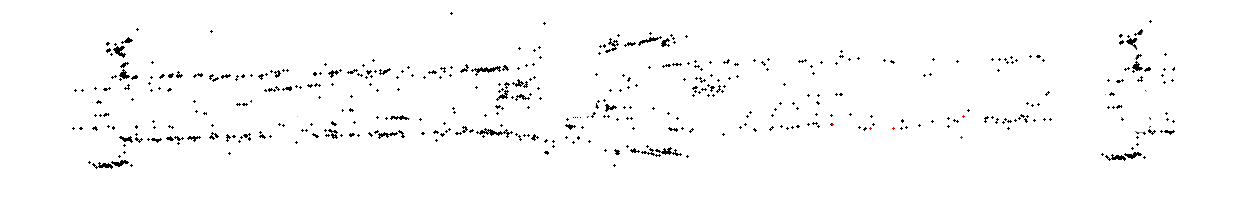}
\caption{Mapping using ORB-SLAM3}
\label{fig:orb_map}
\end{figure}

\begin{figure}[!t]
\centering
\includegraphics[width=\columnwidth]{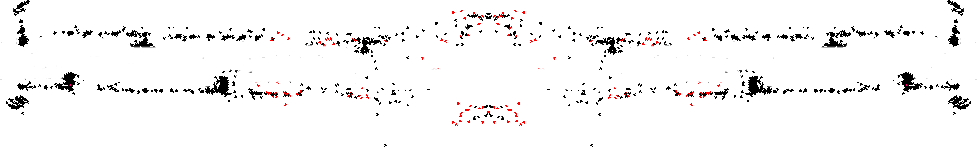}
\caption{Mapping using UcoSLAM}
\label{fig:ucoslam_map}
\end{figure}

\begin{figure}[!t]
\centering
\includegraphics[width=\columnwidth]{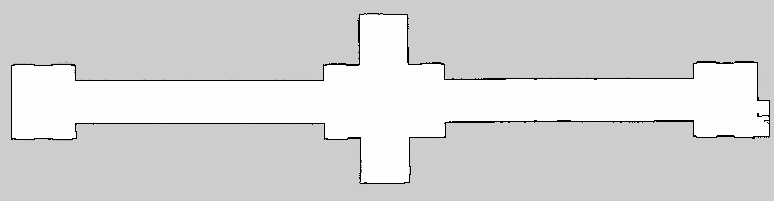}
\caption{Mapping using marker tracker}
\label{fig:marker_map}
\end{figure}

\begin{figure}[!t]
\centering
\includegraphics[width=\columnwidth]{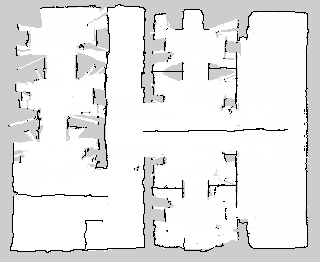}
\caption{Mapping of lab environment using Hector SLAM}
\label{fig:dt_hector_map}
\end{figure}

\begin{figure}[!t]
\centering
\includegraphics[width=\columnwidth]{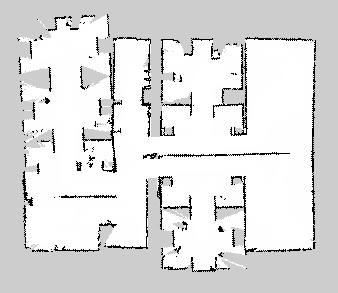}
\caption{Mapping of lab environment using marker tracker}
\label{fig:dt_marker_map}
\end{figure}

\noindent Mapping using Hector SLAM is shown in Figure \ref{fig:hector_map} and \ref{fig:dt_hector_map} and mapping from the proposed method is given in Figure \ref{fig:marker_map} and \ref{fig:dt_marker_map}. Both the mapping methods are compared to the ground truth map using the procedure specified in \cite{Yagfarov2018}. All the occupancy grid maps undergo thresholding so that any cell is either occupied (1) or unoccupied (0), thus converting unknown cells (-1) to unoccupied cells. Next, the thresholded map undergoes thinning such that the obstacles identified are only one pixel wide. Mapping results for ORB-SLAM3 and UCOSLAM are shown in Figures \ref{fig:orb_map} and \ref{fig:ucoslam_map} respectively. Since both ORB-SLAM3 and UCOSLAM are feature based methods, the grid map obtained is not dense. Obtained results are aligned with the ground truth using Iterative Closest Point (ICP) \cite{May2009} and Average Distance to Nearest Neighbor (ADNN) \cite{Yagfarov2018} is computed for all obstacle cells across the approaches used. The paper \cite{Yagfarov2018} calculates the error in centimeters, but we use pixel-wise distances as conversion from occupancy grid cells to distances is simply multiplication by a scaling factor (5cm per pixel). ADNN-based errors for the tested methods are given in Table \ref{tab:mapping_result}. The table also contains the time taken for mapping using the respective methods.
\begin{table*}
\begin{center}
\caption{Mapping metrics.}
\label{tab:mapping_result}
\begin{tabular}{| c | c | c | c | c | c |}
\hline
\textbf{Metric}&\textbf{Environment}& \textbf{Hector}& \textbf{ORB}& \textbf{UCO} & \textbf{Our}\\
&& \textbf{SLAM}& \textbf{SLAM}& \textbf{SLAM} & \textbf{Approach}\\
\hline
\multirow{ 2}{*}{\textbf{ADNN}}& Corridor &51.086 &82.37&68.25& 46.37 \\
             & Lab & 85.93 & 115.38& 92.54& 74.58 \\
\hline
\multirow{ 2}{*}{\textbf{RMSE}}& Corridor & 54.92& 88.49&74.34&49.18  \\
             & Lab & 92.45 &132.22&121.98&88.94  \\
\hline
\multirow{ 2}{*}{\textbf{ATE(m)}}& Corridor & 0.52& 0.64&0.54&0.51  \\
             & Lab & 0.78 & 0.92 &0.65&0.59  \\
\hline
\textbf{Time}& Corridor & 151 & 151&151& 149 \\
\textbf{(sec)}& Lab & 285 &285&285& 291 \\
\hline
\end{tabular}
\end{center}
\end{table*}
It is observed that the time taken is nearly the same in the three methods. This is because the robot is operated at low speed while covering the environment and the map is derived from recorded data from that run. Our method is only slightly slower due to the map completion task of saving the maps to the server. We get better ADNN for our method owing to better localization. ATE is also better due to the localization and the precise motion that can be achieved by the autonomous robot.


\subsection{Marker Placement Results}
\label{sec:plac_res}
\begin{figure*}[!t]
\centering
\includegraphics[width=\textwidth]{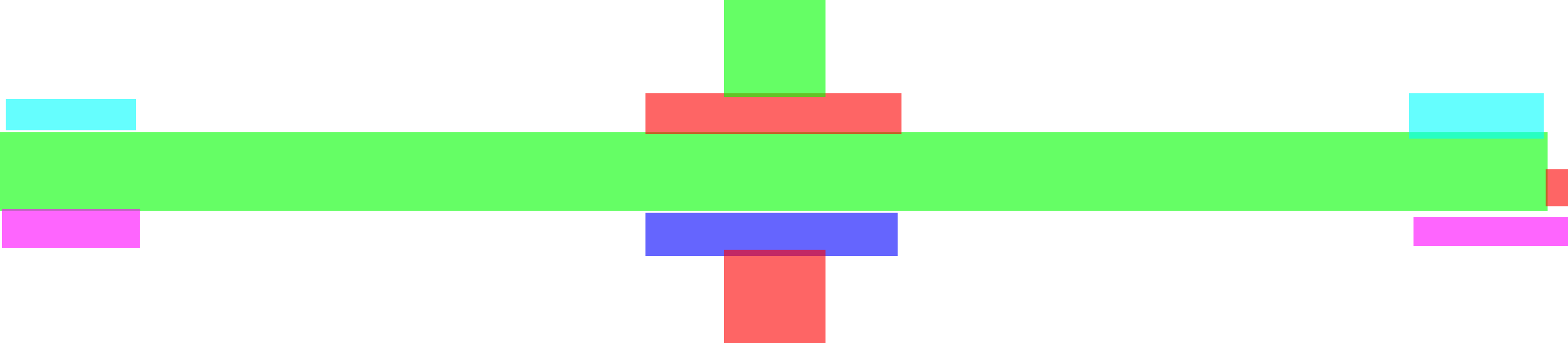}
\caption{Rectangular decomposition of corridor environment}
\label{fig:rectangulate}
\end{figure*}

\begin{figure}[!t]
\centering
\includegraphics[width=\columnwidth]{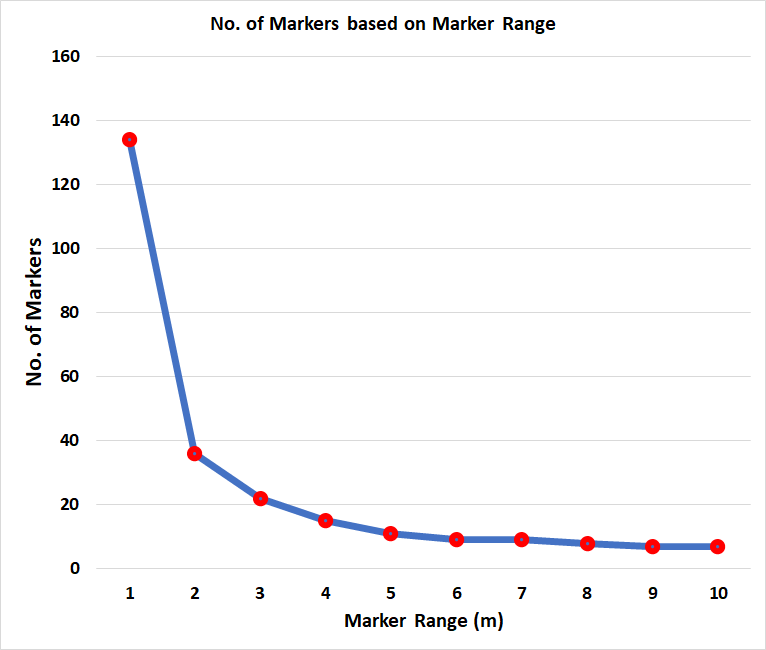}
\caption{Results of reduced marker placement on corridor map over varying ranges}
\label{fig:marker_placement}
\end{figure}

\begin{figure*}[!t]
\centering
\includegraphics[width=\textwidth]{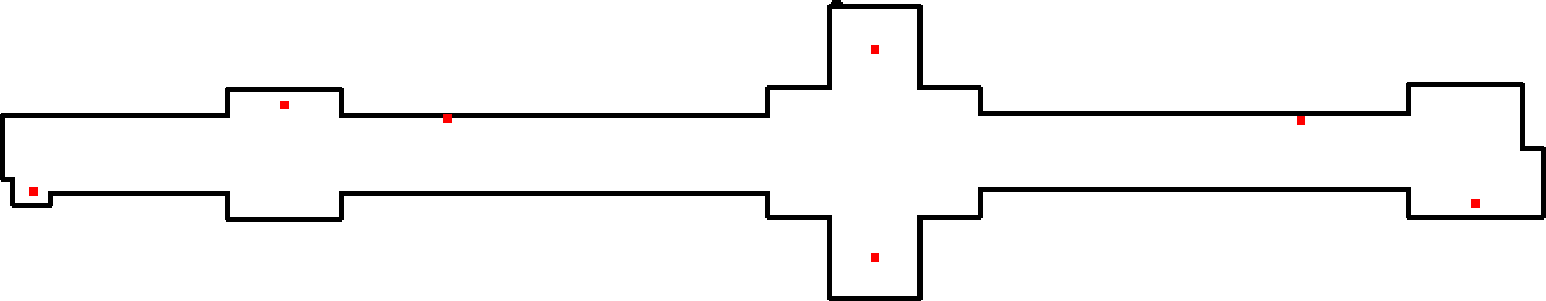}
\caption{Reduced marker placement with 5m range}
\label{fig:placement_5m}
\end{figure*}

\begin{figure}[!t]
\centering
\includegraphics[width=\columnwidth]{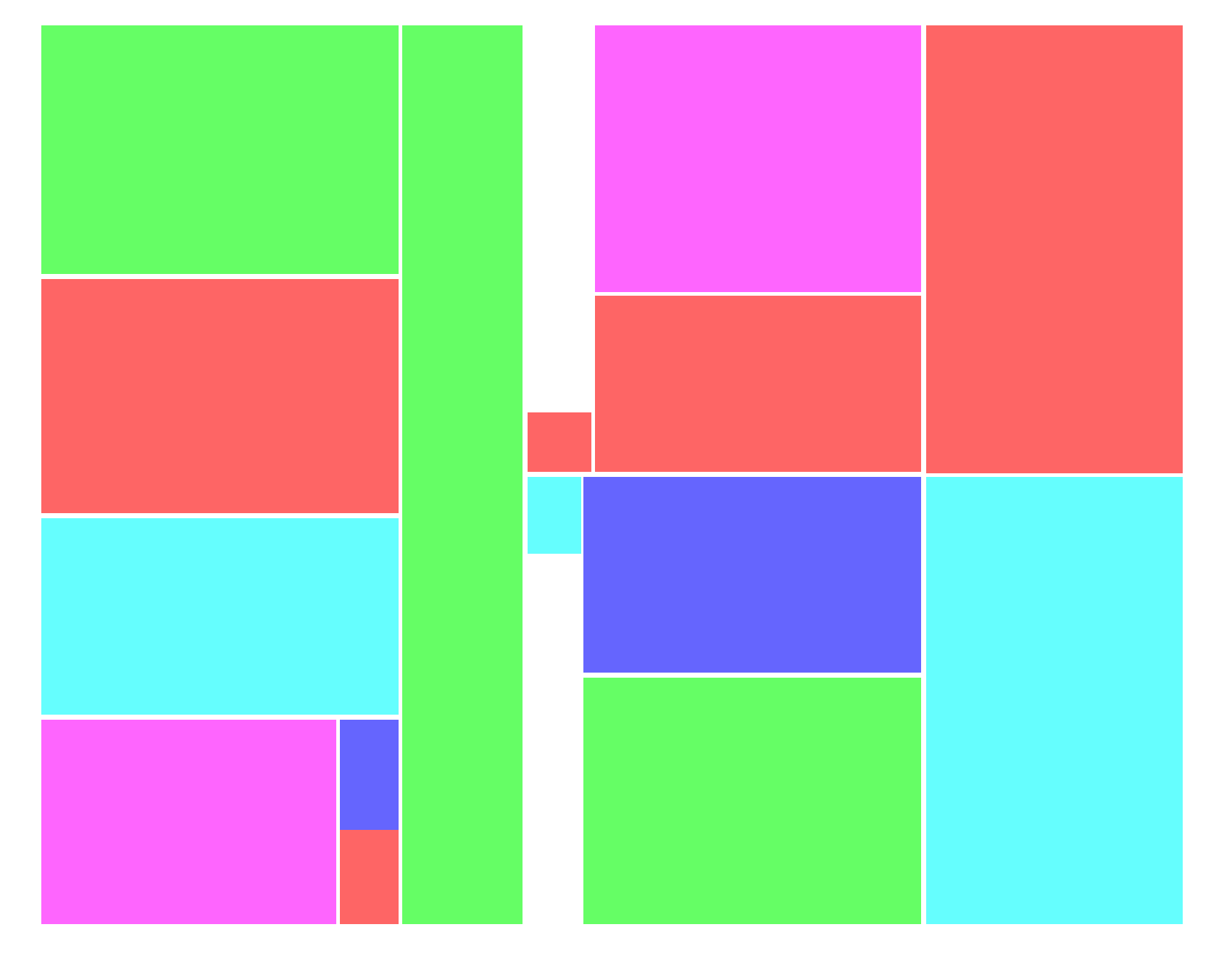}
\caption{Rectangular decomposition of lab environment}
\label{fig:dt_lab_rectangulation}
\end{figure}

\begin{figure}[!t]
\centering
\includegraphics[width=\columnwidth]{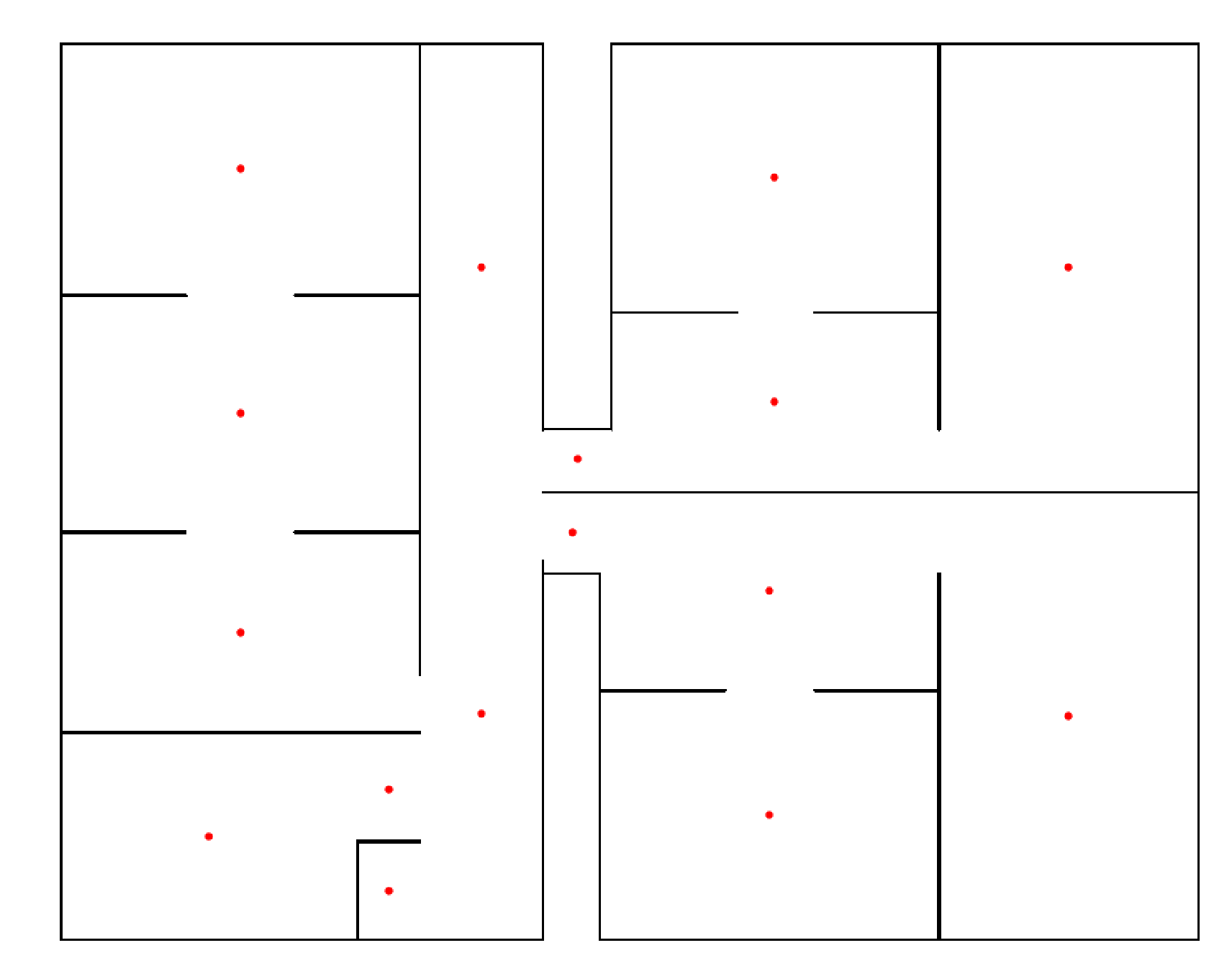}
\caption{Reduced marker placement with 5m range for lab environment}
\label{fig:dt_lab_placement}
\end{figure}

\noindent Given different-sized markers with varying trackable and cutoff ranges, marker placement candidates are selectively removed by the algorithm in different ways to provide a set of marker positions for the given map and marker ranges. The number of markers required to cover the environment is calculated for different marker ranges corresponding to the different marker sizes tested. Rectangular decomposition of the mapped area is given in Figures \ref{fig:rectangulate} and \ref{fig:dt_lab_rectangulation}.
\\
\noindent Figure \ref{fig:marker_placement} shows the results of marker placement for different-sized markers. Different-sized markers offer different ranges. The y-axis represents the number of markers required to ensure 100\% coverage of the environment. For markers with a range of 1 meter, we observe a requirement of 134 markers to cover the given environment. This is because the corridor is 2.16 meters wide and the given marker range makes the algorithm create two rectangles to cover the width of the corridor. From range 2 meters onwards, the graph looks normal. The graph flattens out after the 5 meters range and stays so till 7 meters. Hence, we choose the 4.25 meters range marker with a marker size of 20 cm as it gives the minimum number of markers for the smallest marker size. The reduced marker placement for using markers with a 5 meters range is shown in Figure \ref{fig:placement_5m}. Similarly, reduced marker placement for lab environment is shown in Figure \ref{fig:dt_lab_placement}.

\subsection{Navigation using Reduced Markers in Environment}
\noindent Given the reduced markers placed in the environment, we have ensured complete coverage of the navigable area. Hence, the system is capable of localization to enable navigation. In this section, we evaluate localization performance and navigation performance in the context of a mobile robot as well as a sighted blindfolded participant with the helmet-mounted PT camera.
\subsubsection{Verification of Localization}
\begin{figure}[!t]
\centering
\includegraphics[width=\columnwidth]{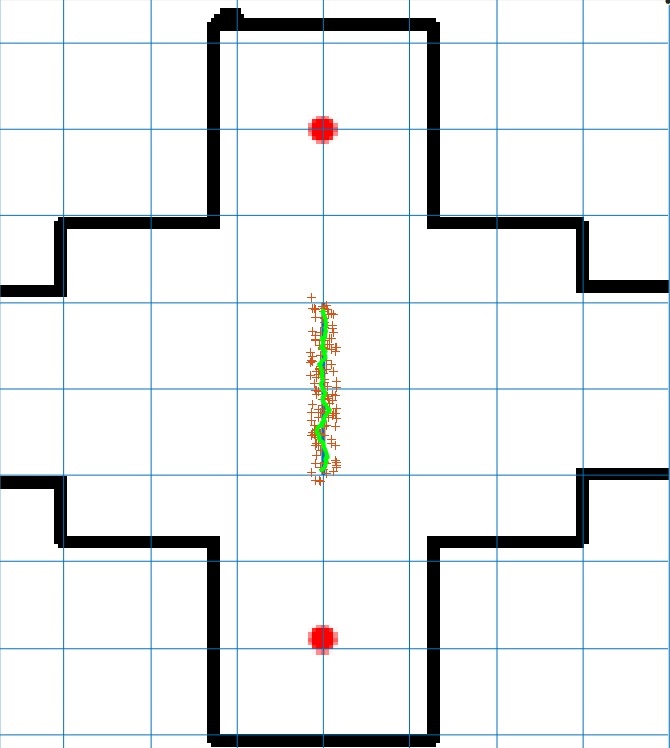}
\caption{Localization of robot when moving on a known linear path. + symbols represent the raw localization data, and the curved green Line represents the smoothed Localization data}
\label{fig:localization}
\end{figure}

Given the reduced markers placed in the environment, it is essential to determine whether the localization information offered is reliable during the motion of the robot. A continuous source of localization information is required for the pure pursuit controller used for navigation. For this experiment, a known set of map points representing a simple linear path is selected. The path is such that it lies between two markers to test the transition feature of the marker tracker system. The robot is driven on the path manually using remote control. The known path and the path estimated by the marker tracker while the robot is in motion are shown in Figure \ref{fig:localization}.
The path is the straight line between the two markers. It can be observed that there is some disturbance in the localization data (plus symbols) when the robot is in motion, owing to pixel level changes in the detection of the marker, reflecting an error of $ \pm 15 cm $.  The error is smoothed out by a simple moving average filter (green line), bringing down the error to within $ \pm 8 cm $.
In contrast, localization based on maps generated by ORB-SLAM3 and UcoSLAM are bound to $\pm1cm$ and $\pm10cm$, respectively.

\subsubsection{Verification of Navigation}
Here we verify if the robot is able to navigate a given path from one section of the environment to another, transitioning between different markers as well as traversing multiple waypoints along the given path. Localization information as the robot is moving along the path is plotted and given as feedback to the pure pursuit controller to calculate deviation from the path and give correction commands to the motion controller. The distance between the path specified for traversal and the actual path traveled by the robot will help estimate the accuracy of the navigation controller and the correction inputs given.
\begin{figure}[!t]
\centering
\includegraphics[width=\columnwidth]{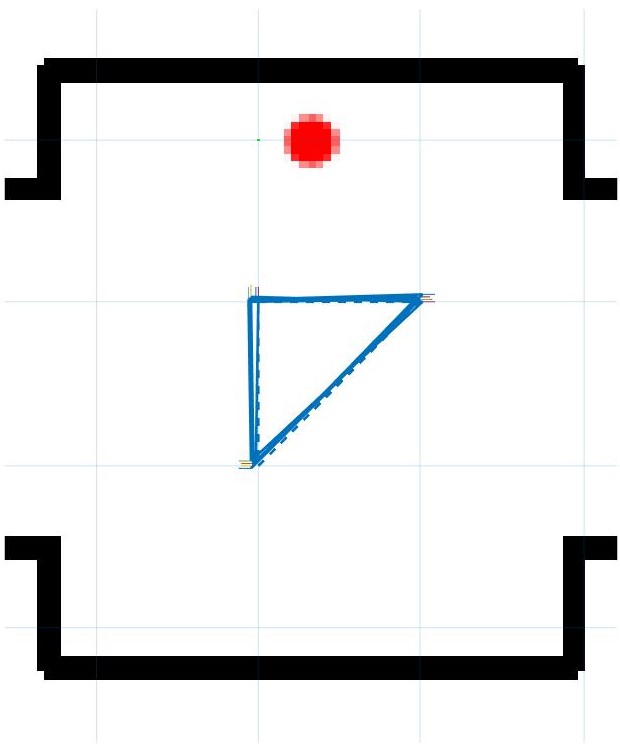}
\caption{Navigation of the robot using location from marker tracker. The dotted line represents the ideal direct path of traversal, and the solid blue lines represent the actual path traversed by the robot}
\label{fig:navigation}
\end{figure}
\\
\noindent Figure \ref{fig:navigation} shows the navigation experiment conducted on a section of the environment. Three waypoints with different orientations were selected, and the robot was made to autonomously cover the waypoints repeatedly. The trajectory followed by the robot is shown in blue, and the waypoints reached are represented as small arrows. It was observed that the error was bounded and within $ \pm 10 cm$.

\subsubsection{Verification of Path Planning and Navigation}
This section describes the experiments conducted to verify the functioning of the complete implementation of the work presented in this paper. In addition to the previous section, the waypoints are presented to the system dynamically by simulated user input, and the path planning and navigation stack using the marker tracker are put to the test. 

\section{Conclusion}
\noindent We have developed a VA system to assist the navigation of BVI individuals in indoor environments. The marker-based localization used here is accurate up to $\pm8cm$. The PT turret-mounted camera, in combination with planar markers mounted on a cuboid, ensures that a marker can be tracked from all directions. This helps us to cover a large environment with significantly fewer markers and ensures minimal defacing. The proposed marker placement makes good use of the improved FOV and reduces the number of markers to cover the environment. Placing markers in the environment is also cost-effective and scalable. We map the environment using localization information from the markers and range information from a laser scanner. Localization and mapping results are better in comparison to state-of-the-art methods like Hector-SLAM, ORB-SLAM, and UcoSLAM in terms of RMSE, ADNN, and ATE. Our method performs significantly better in comparison to ORB-SLAM, and UcoSLAM. However, we observed marginal improvement over Hector-SLAM.
To address changes in layout or features, remapping of the environment will be required. Implementation of a partial mapping of the environment can reduce the efforts involved.
\noindent Proposed extensions to this work include autonomous exploration of the unknown terrain. Autonomous exploration by the robot would further reduce the human effort required to set up the VA system. This system can be extended to multi-floor layouts by repeating the mapping process for each floor and adding height information to the known positions of the markers. The addition of semantic features to the environment can provide a rich dictionary of destinations in the map. We will extend the mapping system to include semantic features and develop the audio-based user interface for the BVI user.

\backmatter

\bmhead{Acknowledgments}

The Authors would like to acknowledge the students of INSPIRE-LAB, Department of Computer Science, BITS Pilani, for helping with the experiments. We would also like to thank the Department of Computer Science, BITS Pilani, for providing resources and access to sections of the department for mapping.

\section*{Statements and Declarations}

\begin{itemize}
\item Funding: 
Dharmateja Adapa is supported by the Prime Ministers Research Fellowship undertaken by the Science and Engineering Research Board (SERB) and the Federation of Indian Chambers of Commerce and Industry (FICCI) in association with Industry Partner AutoNxt Automation Pvt. Ltd.

\item Competing interests: 
The authors have no competing interests to declare that are relevant to the content of this article.

\item Ethics approval: 
Not applicable

\item Consent to participate: 
Not applicable

\item Consent for publication: 
Not applicable

\item Availability of data and materials: 
Data sharing not applicable to this article as no datasets were generated or analysed during the current study.

\item Code availability: 
Not applicable

\item Authors' contributions: 
All authors contributed to the study, conception, and design. Material preparation, data collection, and analysis were performed by Dharmateja Adapa, Virendra Singh Shekhawat, and Avinash Gautam. The first draft of the manuscript was written by Dharmateja Adapa and all authors reviewed and commented on previous versions of the manuscript. Verification of the methodology used and results obtained were done by Sudeept Mohan. All authors read and approved the final manuscript.

\end{itemize}


\bibliography{BVI_Mapping}


\end{document}